\documentclass[journal]{IEEEtran}

\usepackage{xcolor,soul,framed} 
\usepackage{cite}
\colorlet{shadecolor}{yellow}
\usepackage[pdftex]{graphicx}
\usepackage{float} 
\usepackage{subfigure} 
\graphicspath{{../pdf/}{../jpeg/}}
\DeclareGraphicsExtensions{.pdf,.jpeg,.png}

\usepackage[cmex10]{amsmath}
\usepackage{array}
\usepackage{mdwmath}
\usepackage{mdwtab}
\usepackage{eqparbox}
\usepackage{url}
\usepackage{caption}
\usepackage{amsfonts,amssymb} 
\usepackage{stfloats}
\usepackage{booktabs}
\usepackage{threeparttable}
\usepackage{multirow}

\begin{document}
\bstctlcite{}
\title{DASNet: Dual attentive fully convolutional siamese networks for change detection in high-resolution satellite images}
\author{Jie~Chen,
	Ziyang~Yuan,
	Jian~Peng,
	Li~Chen,
	Haozhe~Huang,
	Jiawei~Zhu,
	Yu Liu\^*,
	and~Haifeng~Li\^*,~\IEEEmembership{Member,~IEEE}
	
 \thanks{This work was supported by the National Natural Science Foundation of China (41671357, 41871364, 41871276, and 41871302). This work was carried out in part using computing resources at the High Performance Computing Platform of Central South University.}
 
\thanks{J. Chen, Z. Yuan, J. Peng, L. Chen, H. Huang, J. Zhu and~H.~Li are with the School of Geosciences and Info-Physics, Central South University (e-mail:lihaifeng@csu.edu.cn).}
\thanks{Y. Liu is with the Department of Systems Engineering, National University of Defense Technology. Yu Liu and Haifeng Li are the corresponding authors}  

\thanks{Citation: Jie Chen, Ziyang Yuan, Jian Peng, Li Chen, Haozhe Huang, Jiawei Zhu, Yu Liu, Haifeng Li. DASNet: Dual attentive fully convolutional siamese networks for change detection of high resolution satellite images. IEEE Journal of Selected Topics in Applied Earth Observations and Remote Sensing. 2020. DOI: 10.1109/JSTARS.2020.3037893. }}	
	
\markboth{}
{ \MakeLowercase{\textit{et al.}}: DASNet: Dual attentive fully convolutional siamese networks for change detection in remote sensing images}
	
\maketitle

\begin{abstract}
Change detection is a basic task of remote sensing image processing. The research objective is to identify the change information of interest and filter out the irrelevant change information as interference factors. Recently, the rise in deep learning has provided new tools for change detection, which have yielded impressive results. However, the available methods focus mainly on the difference information between multitemporal remote sensing images and lack robustness to pseudo-change information. To overcome the lack of resistance in current methods to pseudo-changes, in this paper, we propose a new method, namely, dual attentive fully convolutional Siamese networks (DASNet), for change detection in high-resolution images. Through the dual attention mechanism, long-range dependencies are captured to obtain more discriminant feature representations to enhance the recognition performance of the model. Moreover, the imbalanced sample is a serious problem in change detection, i.e., unchanged samples are much more abundant than changed samples, which is one of the main reasons for pseudo-changes. We propose the weighted double-margin contrastive loss to address this problem by punishing attention to unchanged feature pairs and increasing attention to changed feature pairs. The experimental results of our method on the change detection dataset (CDD) and the building change detection dataset (BCDD) demonstrate that compared with other baseline methods, the proposed method realizes maximum improvements of 2.9\% and 4.2\%, respectively, in the F1 score. Our PyTorch implementation is available at https://github.com/lehaifeng/DASNet.

\end{abstract}


%
\IEEEpeerreviewmaketitle


\section{Introduction}
\IEEEPARstart{R}{emote} sensing change detection is a technique for detecting and extracting the change information in a geographical entity or phenomenon through multiple observations \cite{singh1989review}. Remote sensing image change detection research plays a crucial role in land use and land cover change analysis, forest and vegetation change monitoring, ecosystem monitoring, urban expansion research, resource management, and damage assessment \cite{xian2009updating,8306282,7464251,coppin2004review,luo2018urban,lu2004change,brunner2010earthquake,zelinski2014use}. With the development of satellite imaging technology, many high-resolution remote sensing images can be obtained more easily. In high-resolution remote sensing images, ground objects have more space and shape features; hence, high-resolution remote sensing images become an important data source for change detection \cite{bruzzone2012novel}. The effective extraction and learning of the rich feature information in high-scoring remote sensing images, reduction in the interference of pseudo-changes, which means changes that are not truly occurring and changes that do not interest us, and further improvement in the accuracy of change detection are important issues in the field of remote sensing change detection.

Traditional change detection methods can be divided into two categories according to the research objects: pixel-based change detection methods and object-based change detection methods \cite{hussain2013change}. Pixel-based change detection methods typically generate a difference graph by directly comparing the spectral information or texture information of the pixels and obtain the final result graph via threshold segmentation or clustering \cite{bruzzone2000automatic,celik2009unsupervised,deng2008pca,wu2017post,huang2008use,cao2014automatic,volpi2013supervised}. Although this method is simple to implement, it ignores the spatial context information and generates a substantial amount of salt-and-pepper noise during processing. Another type of method divides the remote sensing image into disjoint objects and uses the rich spectral, textural, structural, and geometric information in the image to analyze the differences of temporal images \cite{zhang2017object,8335352,gil2016description,qin2013object,ma2016object}. Although this type of method uses the spatial context information of high-resolution remote sensing images, traditional manual feature extraction is complex and exhibits poor robustness.

In recent years, deep learning has yielded impressive results in image analysis, natural language processing, and other fields \cite{lecun2015deep,7406686,8409959}. Neural networks with FCN \cite{long2015fully} structures are widely used in remote sensing change detection tasks \cite{liu2016deep,zhan2017change,alcantarilla2018street,daudt2018fully,papadomanolaki2019detecting,8736213}. These methods can be divided into two categories: methods in the first category fuse unchanged images and changed images and input them into a network with FCN structure to detect the changes by maximizing the boundaries instead of directly measuring the changes. The methods in the other category detect changes by measuring the distances between feature pairs that are extracted from images.

However, the available methods have low robustness to pseudo-changes. There are two main reasons for this:

First, due to the lack of satisfactory features that can effectively distinguish between changed areas and unchanged areas, the available features are often sensitive to factors such as noise, angle, shadow, and context.

Second, the distributions of the changed and unchanged data are often severely unbalanced, namely, the number of unchanged samples is much larger than the number of changed samples.

To address the first problem, we propose the DASNet framework. The basic strategy is to use a dual attention mechanism to locate the changed areas and to obtain more discriminant feature representations, which renders the learned features more robust to changes. Many previous studies \cite{byeon2015scene,shuai2017scene} have demonstrated that more distinguishable feature representations can be obtained by capturing long-range dependencies \cite{7940028}. Attention mechanisms can model long-range dependencies and have been widely used in many tasks \cite{lin2016efficient,lin2017structured,shen2018disan,vaswani2017attention,tang2011image,tang2015rgb}. Among them, the self-attention mechanism \cite{wang2018non,zhang2018self} explores the performances of nonlocal operations of images and videos in the space-time dimension. The self-attention mechanism is of substantial significance for the long-range dependencies of images. Additionally, many researchers\cite{8100150,Woo_2018_ECCV,fu2019dual} began to combine spatial attention and channel attention so that the network could not only focus on the region of interest but also improve the discrimination of features. Therefore, we introduce an extended attention mechanism \cite{fu2019dual}, which is composed of a channel attention module and a spatial attention module, for obtaining more discriminative image features to enhance the model's performance in recognizing changes and to improve its robustness to pseudo-changes.

Aiming at overcoming the second problem, due to the imbalance of data in change detection and the unbalanced contributions to the network of the changed feature pairs and the unchanged feature pairs in the contrastive loss \cite{hadsell2006dimensionality} of traditional Siamese networks, we established with weighted double-margin contrastive (WDMC) loss. The WDMC loss can mitigate the effects of having more unchanged regions than changed regions in the original data by setting weight coefficients. It can also alleviate the imbalance in the punishment between the unchanged feature pairs and the changed feature pairs during network training by setting double margins.

The main contributions of this paper are as follows:

1. We propose a new remote sensing change detection method that is based on deep metric learning, which uses dual attention modules to improve feature discrimination to more robustly distinguish changes.

2. We propose the weighted double-margin contrastive loss (WDMC loss), which increases the distance between changed feature pairs and reduces the distance between unchanged feature pairs while balancing the impacts of changed regions and unchanged regions on the network. Thereby enhancing the network's performance in recognizing change information and its robustness to pseudo-changes..

3. Compared with the selected baselines, the proposed method yielded SOTA results on the CDD dataset \cite{lebedev2018change} and the BCDD dataset \cite{ji2018fully}, and the F1 scores reached 91.9\% and 89.8\%, respectively. Compared with other baseline methods, the maximum increases were 2.9\% and 4.2\%, respectively.

The remainder of the paper is organized as follows. Section II reviews the related works. Section III describes the proposed method in detail. To evaluate our method, experiments are designed in Section IV, and the proposed method is discussed. Finally, Section V summarizes our work in this paper.

\section{Related works}

Change detection is a basic task in the field of remote sensing, and researchers have developed many change detection technologies for this task. Remote sensing change detection methods typically include feature extraction and changing area identification. The objective of the former is to extract meaningful features, such as color distribution, texture characteristics, and context information. The objective of the latter is to use technical algorithms to analyze previously extracted features to identify the changing regions in multitemporal remote sensing images. Based on the techniques that are utilized in these change detection methods, we divide the remote sensing change detection methods into traditional change detection methods and deep learning change detection methods.

Traditional change detection methods mainly accept the feature differences and ratios of pairs of pixels or objects as input and detect changes by determining thresholds. Change vector analysis (CVA) \cite{bruzzone2000automatic} conducts different calculations on the data of each band of images in various periods to obtain the change amount in each pixel in each band to form a change vector, and it identifies the changed and unchanged regions using thresholds. Principal component analysis (PCA) \cite{deng2008pca}, which is a classical mathematical transformation method, obtains the difference image by extracting the effective information in the image bands, performs the difference calculation with the first principal component and obtains the change map via threshold segmentation. Multivariate alteration detection (MAD) \cite{nielsen1998multivariate} extracts change information by maximizing the variance in the image difference. Slow feature analysis (SFA) \cite{wu2013slow} can extract invariant features from multitemporal remote sensing image data, reduce the radiation differences of unchanged pixels, and improve the separability between changed and unchanged pixels.

The remote sensing change detection methods that are based on deep learning use the features that are extracted from multitemporal images by deep neural networks to determine the change information in ground objects. In the field of natural image change detection, deep learning methods perform well \cite{guo2018learning,varghese2018changenet}. In the field of remote sensing, the strategy of using deep learning for change detection has also been utilized. SCCN \cite{liu2016deep} uses a deep symmetrical network to study changes in remote sensing images, and DSCN \cite{zhan2017change} uses two branch networks that share weights for feature extraction and uses the features that are obtained by the last layer of the two branches for threshold segmentation to obtain a binary change map. CDNet \cite{alcantarilla2018street}, FC-EF, FC-Siam-Conc, FC-Siam-Diff in \cite{daudt2018fully}, and BiDateNet \cite{papadomanolaki2019detecting} implement end-to-end training on the change detection dataset and learn the decision boundary to obtain the change map.

In contrast to the above methods, the proposed method directly measures changes and obtains a more discriminative feature representation through the dual attention module, which can capture long-range dependencies. Our proposed method uses the WDMC loss to improve the degree of intraclass compactness and to balance the influences of the changed regions and the unchanged regions on the network to enhance the recognition performance of the network for the changed information.

\section{Methodology}

\begin{figure*}[htb]
  \begin{center}
  \includegraphics[width=1\textwidth]{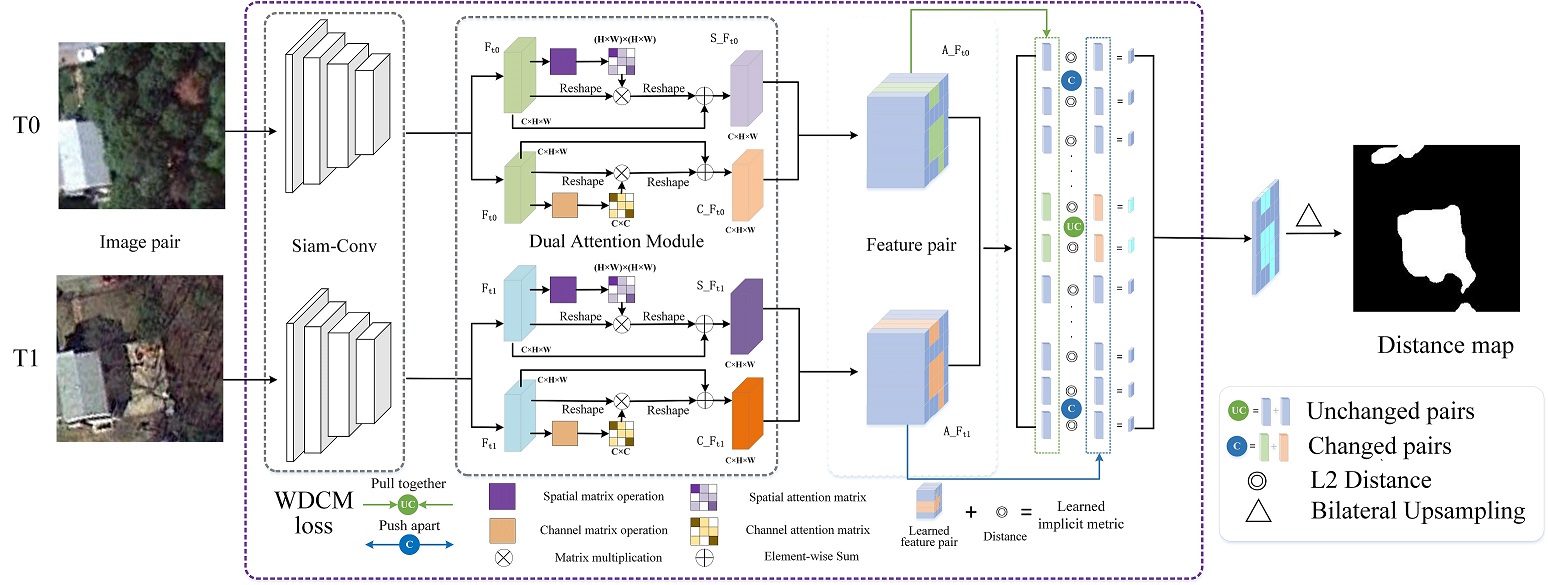}
  \caption{Overview of the dual attentive fully convolutional Siamese network.}\label{p1}
  \end{center}
\end{figure*}

In this section, we introduce the proposed network in detail. First, the overall structure of the network that is proposed in this paper is introduced (Fig.~\ref{p1}). The spatial attention and channel attention modules are explained subsequently. Finally, the proposed WDMC loss function is introduced.

\subsection {Overview}
Compared with general optical remote sensing images, high-resolution remote sensing images have more abundant information and higher requirements for feature extraction. We first use the Siam-Conv module to generate local features ($F_{t0}$, $F_{t1}$) by inputting high-resolution image pairs that are obtained at different times in the same region. Then, the dual attention mechanism is used to establish the connections between local features, which are used to obtain global context information to better distinguish the changing area from the unchanged area.

Consider spatial attention as an example, which generates new features that contain spatial long-range contextual information via the following three steps. The first step is to use a spatial attention matrix to model the spatial relationship between any two pixels of the features obtained from Siam-Conv. The next step is to conduct matrix multiplication between the spatial attention matrix and the original features. Third, the final representations are obtained by applying the element-by-element summation operation on the matrix that was obtained in the second step, and the original features. The channel attention module captures the long-range context information in the channel dimension.

The process of capturing channel relationships is similar to that of spatial attention modules. The first step is to compute the channel attention matrix in the channel dimension. The second step is to conduct matrix multiplication between the channel attentional matrix and the original features. Finally, the matrix that is obtained in the second step and the original features are summed element-by-element. After that, the outputs of the two attention modules are aggregated to obtain a better representation of the features.

Then, the features A\_$F_{t0}$ and A\_$F_{t1}$ that are obtained through the dual attention mechanism are mapped to the feature space. We use the WDMC loss function to decrease the pixel pair distance of the unchanged area and to increase the pixel pair distance of the changing area. Our method uses metric learning and uses the distance between deep features as the basis for discrimination, thereby substantially increasing the accuracy of change detection.

\subsection {Spatial Attention Mechanism}

\begin{figure}[H]
  \begin{center}
  \includegraphics[width=2.8in]{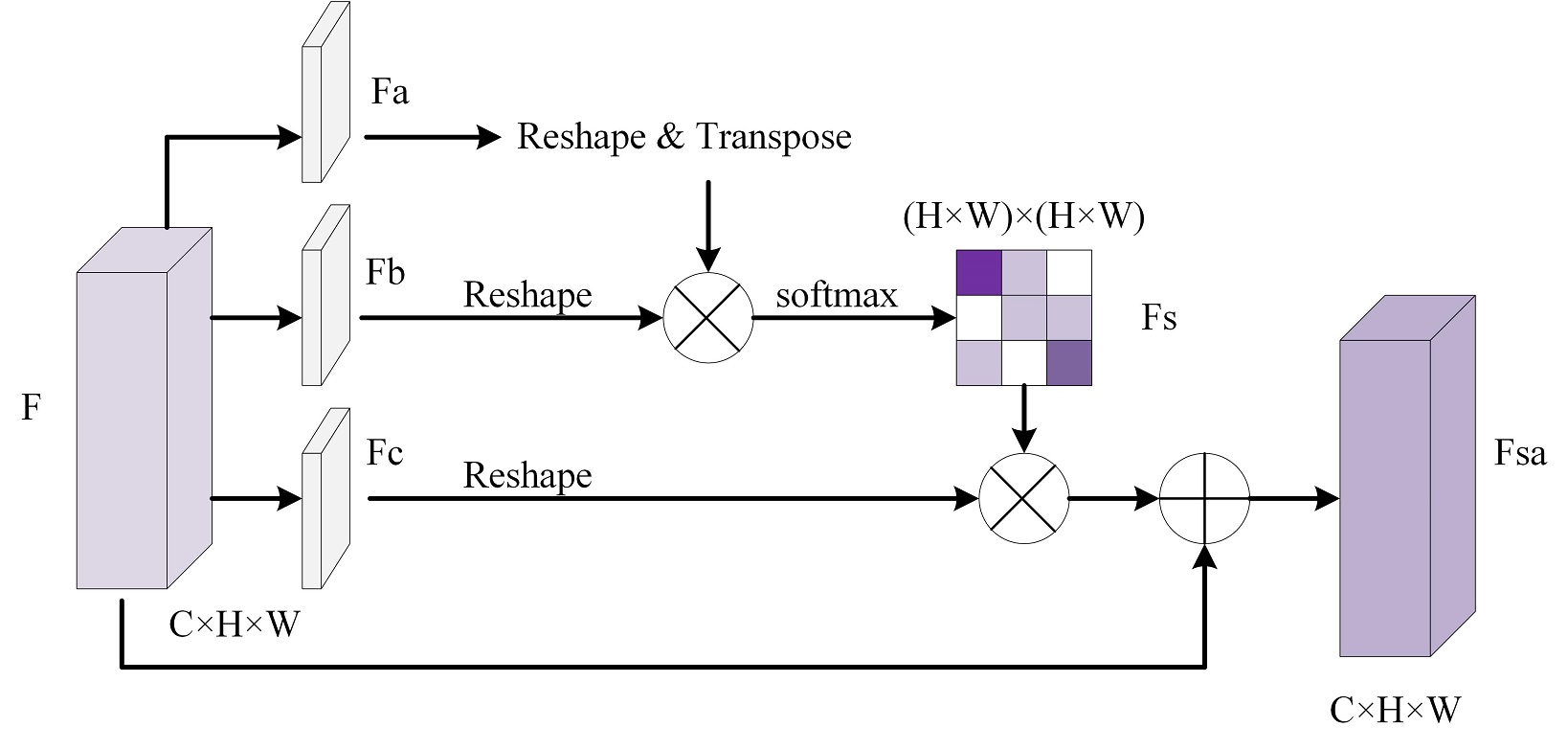}
  \caption{Spatial attention mechanism.}\label{p2}
  \end{center}
\end{figure}

As discussed previously, discriminant feature representations are important for determining the types of features and, thus, for identifying changed and unchanged areas. However, many studies \cite{peng2017large,zhao2017pyramid} have shown that using only local features that are generated by traditional FCNs may lead to misclassification. To model the rich context of local features, we introduce a spatial attention module. The spatial attention module encodes the context information of a long range into local features, thereby enhancing the feature representations.

As illustrated in Fig.~\ref{p2}, the feature F $\in \mathbb{R}^{C \times H \times W}$, where C represents the number of channels of feature F, while H and W represent the height and width of feature F, respectively, that is obtained by Siam-Conv is input into 3 convolutional layers that have the same structure to obtain 3 new features, namely, Fa, Fb and Fc, where \{Fa, Fb, Fc\} $\in \mathbb{R}^{C \times H \times W}$. Then, we reshape Fa and Fb to $\mathbb{R}^{C \times N}$, where $ N = H \times W$. After that, we conduct matrix multiplication between the transpose of Fb and Fa and obtain the spatial attention map Fs $ \in \mathbb{R}^{N \times N}$ through a softmax layer.

\begin{equation}\label{sam1}
Fs_{ji} = 
 \frac{exp (Fa_{i}  \cdot  Fb_{j})}{\sum_{i=1}^{N}{exp (Fa_{i} \cdot Fb_{j})}}\qquad
\end{equation}

$Fs_{ji}$ can be used to measure the effectiveness of the feature at position i on the feature at position j. The stronger the connection between the two features, the larger the value of $Fs_{ji}$.

We reshape Fc to $\mathbb{R}^{C \times N}$ and conduct matrix multiplication with Fs to obtain the result and reshape it to $\mathbb{R}^{C \times H \times W}$. Finally, we multiply the result from the previous step by a scale parameter $\eta$ and perform an elementwise summation operation with F to obtain the final output:

\begin{equation}\label{sam2}
Fsa_{j} = \eta \sum_{i=1}^{N}{(Fs_{ji}Fc_{j}) + F_{j}}
\end{equation}

where $\eta$ is initialized as 0 and gradually learns to assign more weight. From formula (\ref{sam2}), it can be concluded that the resulting feature Fsa at each position is the result of a weighted sum of the features at all positions and the original features. Therefore, it has a global context view and selectively aggregates contexts based on spatial attention maps. Similar semantic features promote each other, which improves the compactness and semantic consistency within the class and enables the network to better distinguish between changes and pseudo-changes.

\subsection{Channel Attention Mechanism}

\begin{figure}[H]
  \begin{center}
  \includegraphics[width=2.8in]{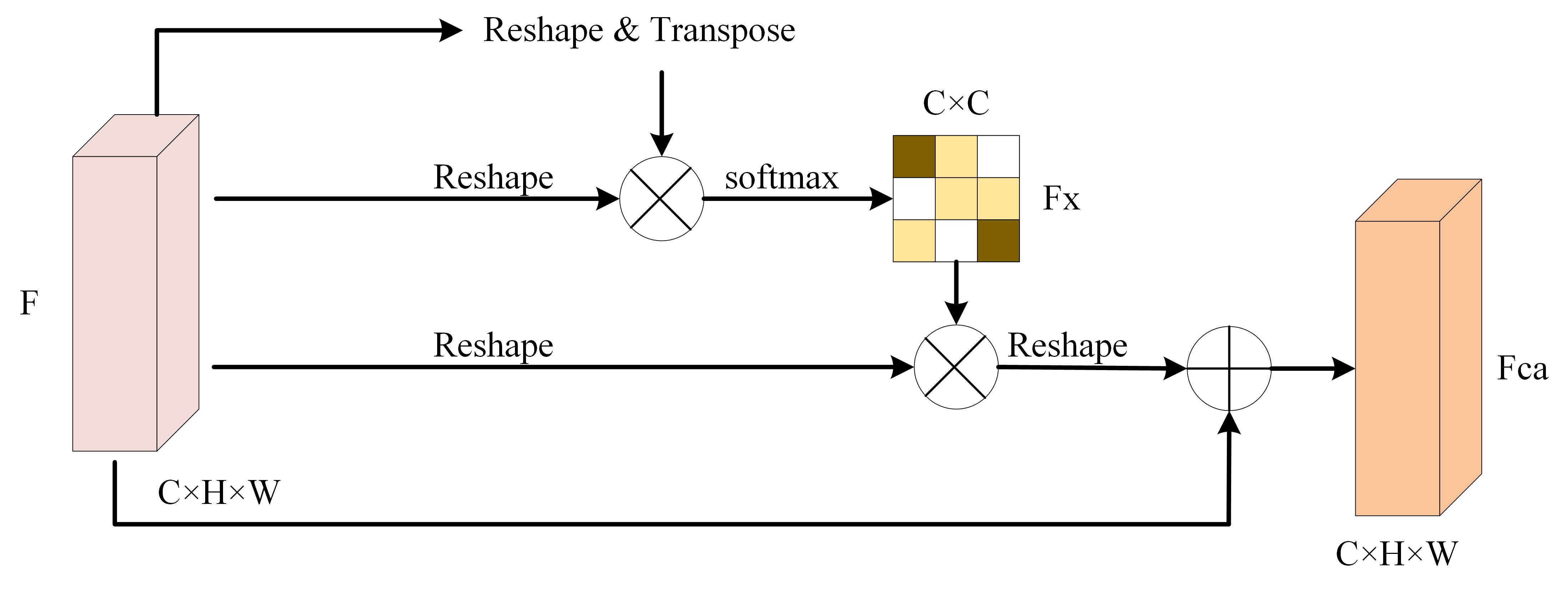}
  \caption{Spatial attention mechanism.}\label{p3}
  \end{center}
\end{figure}

Each high-level feature channel map can be regarded as a response to a ground object, and the semantic responses are related to each other. By utilizing the correlation between channel maps, interdependent feature maps can be enhanced, and feature representations with specified semantics can be improved to better distinguish changes. Therefore, we construct a channel attention module for establishing the relationships between channels.

As illustrated in Fig.~\ref{p3}, in contrast to the spatial attention module, the convolution operation is not used to obtain the new features in the channel attention module. Feature F $\in \mathbb{R}^{C \times H \times W}$ that was obtained by Siam-Conv is reshaped to $\mathbb{R}^{C \times N}$, where $ N = H \times W$. After that, matrix multiplication is performed between the transpose of F and F to obtain the channel attention map Fx $ \in \mathbb{R}^{N \times N}$ through a softmax layer:

\begin{equation}\label{cam1}
Fx_{ji} = 
 \frac{exp (F_{i}  \cdot  F_{j})}{\sum_{i=1}^{C}{exp (F_{i} \cdot F_{j})}}\qquad
\end{equation}

$Fx_{ji}$ can be used to measure the impact of the ith channel on the jth channel. Similarly, the stronger the connection between the two channels, the larger the value of $Fx_{ji}$.

We reshape F to $\mathbb{R}^{C \times N}$ and conduct matrix multiplication with Fx to obtain the result. Finally, we multiply the result from the previous step by a scale parameter $\gamma$ and conduct an elementwise summation operation with F to obtain the final output:

\begin{equation}\label{cam2}
Fca_{j} = \gamma \sum_{i=1}^{C}{(Fc_{ji}F_{i}) + F_{j}}
\end{equation}

where $\gamma$ is initialized as 0 and gradually learns to assign more weight. From formula (\ref{cam2}), it is concluded that the final feature of each channel is the result of a weighted sum of the features of all channels and the original feature, which models the long-term semantic dependency between the feature graphs. It enhances the identifiability of the feature and highlights the feature representation of the region of change.

\subsection{WDMC Loss Function}

In the traditional contrastive loss, a problem of imbalance is encountered in the punishment for the changed feature pairs and the unchanged feature pairs during training. The traditional contrastive loss can be formulated as:

\begin{equation}\label{loss1}
\begin{aligned}
contrastive \;loss =& \sum_{i,j} \,\frac{1}{2}\; [(1-y_{i,j})\;d_{i,j}^{2}\;\\&+\;y_{i,j}\;max\;(d_{i,j}-m,0)^2]
\end{aligned}
\end{equation}

We denote the feature map of the unchanged image through our model as $f_{0}$ and the feature map of the changed image through our model as $f_{1}$, $d_{i,j}$ is the distance between the feature vectors of $f_{0}$ and $f_{1}$ at (i,j), m is the margin that is enforced for changed feature pairs, and y $\in \{0, 1\}$, where y = 0 if the corresponding pixel pair is deemed unchanged, and y = 1 if it is deemed changed.

According to formula (\ref{loss1}), the traditional contrastive loss does not contribute to the loss value only if the distance between the unchanged feature pairs is 0. However, in the remote sensing change detection task, the unchanged area is affected by various imaging conditions, and the imaging difference is sometimes very large; hence, many noise changes will hinder the optimization of the distance between the unchanged feature pairs to 0. However, the changed feature pairs can only contribute to the loss if the distance exceeds the margin, which results in the imbalance of the punishment degree in the training process and affects the network's judgment of the change. In addition, the unchanged area in remote sensing change detection tasks is much larger than the changed area; hence, there is a severe imbalance between the unchanged samples and the changed samples. In response to these problems, we propose the WDMC loss function.

The proposed WDMC loss can be formulated as follows:

\begin{equation}\label{loss2}
\begin{aligned}
WDMC \; loss =& \sum_{i,j} \; \frac{1}{2} \; [w_{1}(1-y_{i,j}) \; max (d_{i,j}-m_{1},0)^{2}\\+&\;w_{2} \; y_{i,j} \; max(d_{i,j}-m_{2},0)^2]
\end{aligned}
\end{equation}

where $m_{1}$ and $m_{2}$ represent the margins of the unchanged sample pairs and the changed sample pairs, respectively, and $w_{1}$ and $w_{2}$ represent the weights of the unchanged feature pairs and the changed feature pairs, respectively.

\begin{equation}\label{loss3}
w_{1} =  \frac{1}{P_{U}} 
\end{equation}

\begin{equation}\label{loss4}
w_{2} =  \frac{1}{P_{C}} 
\end{equation}
where $P_{U}$ and $P_{C}$ are the frequencies of the changed and unchanged pixel pairs, respectively.

According to formula (\ref{loss2}), by setting the margins for the unchanged sample pairs, we can alleviate the imbalance in the punishment between the unchanged feature pairs and the changed feature pairs during network training. We set the weight coefficients to mitigate the effects of having more unchanged regions than changed regions in the original data. Finally, these parameters balance the network's interest in the changed area and the unchanged area.

Inspired by the strategy of deep supervision \cite{lee2015deeply,xie2015holistically}, we calculate the WDMC loss for the feature pairs that are obtained through the spatial attention module, the feature pairs that are obtained by the channel attention module and the final output feature pairs. $\lambda_{i}$ represents the weight for each loss, $L_{sa}$ represents the WDMC loss that is calculated using the feature pairs that are extracted from the spatial attention module, $L_{ca}$ represents the WDMC loss that is calculated using the feature pairs that are extracted from the channel attention module, and $L_{e}$ represents the WDMC loss that is calculated using the final output feature pairs.

Thus, our loss function is ultimately expressed as:
\begin{equation}\label{loss5}
Loss = \lambda_{1}L_{sa}\;+\lambda_{2}L_{ca}+\;\lambda_{3}L_{e}
\end{equation}

In summary, we balance the contributions of the unchanged feature pairs and the changed feature pairs to the loss value during the training process and use the strategy of deep supervision to enhance the feature representation performance of the hidden layers. The discriminative feature representations improve the network's performance in recognizing changes.

\subsection{Implementation Details}

Now, we introduce in detail the relevant parameter design in the model design process.

First, we describe the design of the network structure. In terms of Siam-Conv network structure selection, we chose two basic networks, VGG16 \cite{Simonyan15} and ResNet50 \cite{7780459}. For VGG16, we keep only the first five convolution modules, and we remove the max-pooling layer of the last module. In the first 5 convolution modules, the size of the convolution kernel is.$3\times3$. For ResNet50, we remove the downsampling operations and employ dilated convolutions in the last two ResNet blocks. Then, the dual attention module composed of the spatial attention module and the channel attention module is connected to Siam-Conv to form the complete DSANet.

In terms of loss design, we set four parameters to balance the contributions of the unchanged region feature pairs and the changed region feature pairs to the network and to enhance the network's performance in the identification of changed regions. The values of parameters $w_{1}$ and $w_{2}$ are the pixel ratios of the changed area and the unchanged area in the dataset. The parameters $m_{1}$ and $m_{2}$ must be manually adjusted to optimize the performance of the model.

%

\section{Experiments and Discussion}

To evaluate the performance of the proposed method, we compared other change detection methods on the CDD and BCDD datasets, and we designed ablation experiments for evaluating the proposed structure and improved loss function. Finally, the experimental results are analyzed comprehensively.

\subsection{Databases}
\subsubsection{CDD Dataset}
\begin{figure*}[htb]
  \begin{center}
  \includegraphics[width=0.7\textwidth]{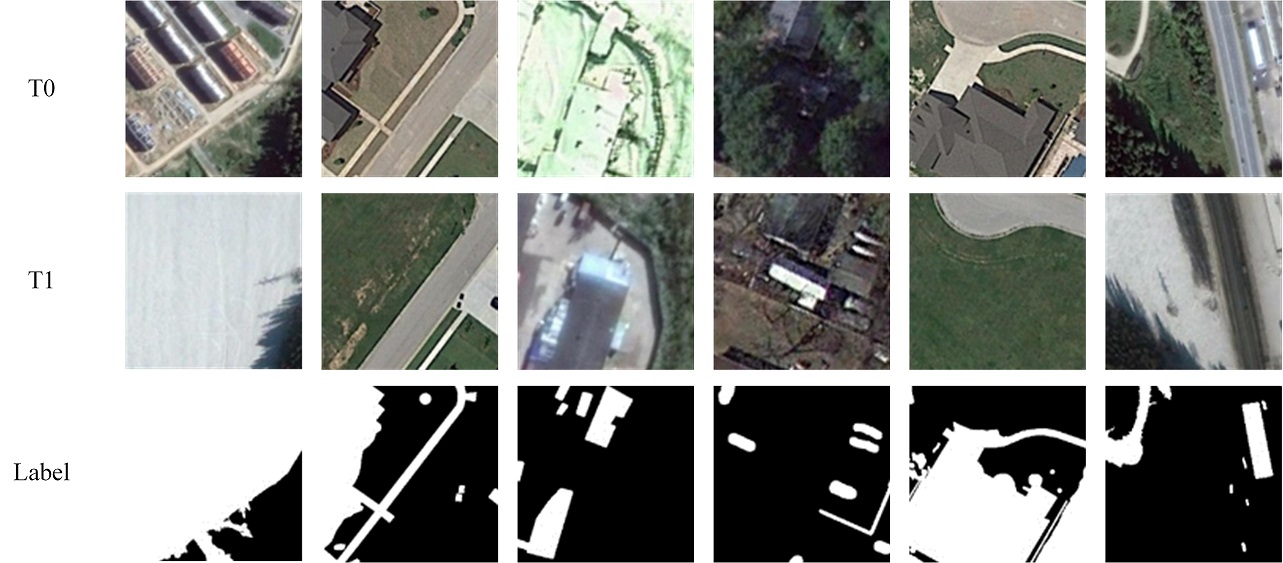}
  \caption{ Multitemporal images that were selected as the training set from the CDD dataset. First row: unchanged images, second row: changed images, and third row: labels.}
  \label{CDD}
  \end{center}
\end{figure*}

CDD (Fig.~\ref{CDD}) is an open remote sensing change detection dataset. The dataset is composed of multisource remote sensing images with 11 original image pairs, which include 7 pairs of seasonal change images with a size of $4,725 \times 2,200$ pixels and 4 pairs of images with a size of $1,900 \times 1,000$ pixels. The resolution varies from 3 cm to 100 cm per pixel, with the seasons varying widely among the bitemporal images. In \cite{ji2018fully}, the author processed the original data to generate datasets with training sets of size 10,000 and test and validation sets of size 3,000 via clipping and rotation.

\subsubsection{BCDD Dataset}

The BCDD dataset (Fig.~\ref{CDD}) covers the area of the 6.3 magnitude earthquake that struck Christchurch, New Zealand, in February 2011. The dataset contains two image scenes that were captured at the same location in 2012 and 2016, along with semantic labels and change detection labels for buildings. Since the size of each image is $32,507 \times 15,354$ pixels, we divided the two images into nonoverlapping $256 \times 256$ -pixel image pairs. Then, we used the cropped images to form the training set, the verification set, and the test set according to the ratio of 8:1:1.

We counted the number of changed pixels and the number of unchanged pixels in the CDD dataset and the BCDD dataset, and the result can be seen in Table \ref{tabdata}. In the CDD dataset, the ratio of changed pixels to unchanged pixels was 0.147. In the BCDD dataset, the ratio was 0.045. This means that in these two datasets, the area that changed is much smaller than the area that has not changed.

\begin{table}
\centering
\caption{Statistics on the number of changed pixels and the number of unchanged pixels in the CDD dataset and BCDD dataset.}
\label{tabdata}
\begin{tabular}{ll|lll}
\hline
\multicolumn{2}{c|}{\textbf{Dataset}} & \multicolumn{1}{c}{\textbf{changed pixels}} & \multicolumn{1}{c}{\textbf{unchanged pixels}} & \textbf{c/uc} \\ \hline
\multirow{4}{*}{CDD}      & train    & 83,981,014                                  & 571,378,986                                   & 0.147                      \\ \cline{2-5} 
                          & val      & 24,676,497                                  & 171,800,431                                   & 0.144                      \\ \cline{2-5} 
                          & test     & 25,411,239                                  & 171,196,761                                   & 0.148                      \\ \cline{2-5} 
                          & total    & 134,068,750                                 & 914,376,178                                   & 0.147                      \\ \hline
\multirow{4}{*}{BCDD}     & train    & 17,071,534                                  & 382,435,922                                   & 0.044                      \\ \cline{2-5} 
                          & val      & 1,854,764                                   & 48,083,668                                    & 0.039                      \\ \cline{2-5} 
                          & test     & 2,426,517                                   & 47,511,915                                    & 0.051                      \\ \cline{2-5} 
                          & total    & 21,352,815                                  & 478,031,505                                   & 0.045                      \\ \hline
\end{tabular}
\end{table}

\begin{figure*}[htb]
  \begin{center}
  \includegraphics[width=0.7\textwidth]{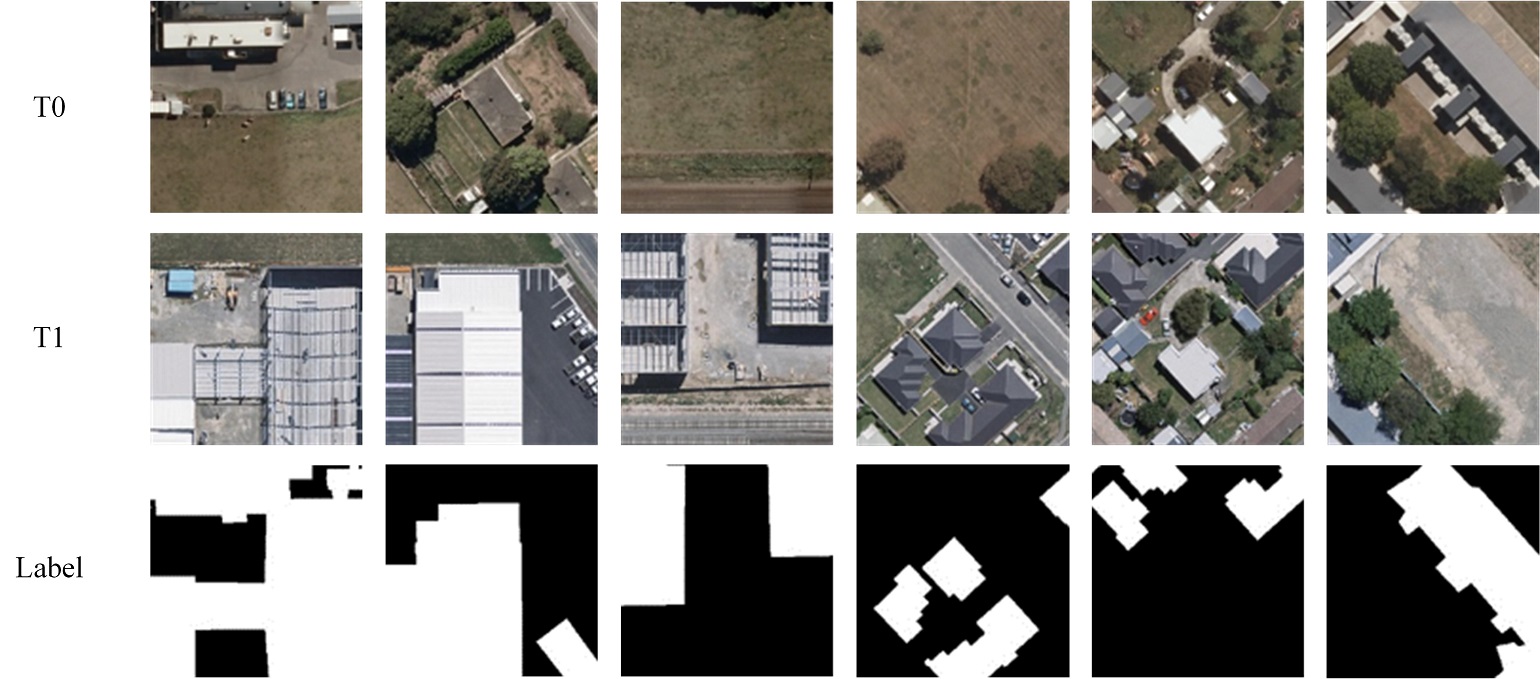}
  \caption{Multitemporal images that were selected as the training set from the BCDD dataset. First row: unchanged images, second row: changed images, and third row: labels.}
  \label{BCDD}
\end{center}
\end{figure*}

\subsection{Metrics and Implementation details}

To evaluate the performance of the proposed method, we used four evaluation indicators: precision (P), recall (R), F1 score (F1), and overall accuracy (OA). In the change detection task, the higher the precision value, the fewer false detections of the predicted result occur, and the larger the recall value, the fewer predicted results were missed. F1 score and OA are the overall evaluation metrics of the prediction results. The larger their values, the better the prediction results. They are expressed as follows:

\begin{equation}\label{pre}
P = \frac{TP}{TP + FP}
\end{equation}

\begin{equation}\label{r}
R = \frac{TP}{TP + FN}
\end{equation}

\begin{equation}\label{F1}
f_{1} = \frac{2PR}{P + R}
\end{equation}

\begin{equation}\label{p}
OA = \frac{TP + TN}{TP + TN + FP + FN}
\end{equation}
where TP is the number of true positives, FP is the number of false positives, TN is the number of true negatives, and FN is the number of false negatives.

The proposed method was implemented with PyTorch as the backend, which is powered by 3 $\times$ GTX TITAN XP. The Adam optimizer was adopted with a learning rate of 1e-4 as an optimization algorithm, and the batch size of the training data was set to 8. According to the experimental performance, the value of $m_{1}$ was set to 0.3, the value of $m_{2}$ was set to 2.2, and the values of $\lambda_{1}$, $\lambda_{3}$ and $\lambda_{2}$ were set to 1.

\subsection{Effect of the WDMC Loss Function}

The measurement values differ between feature pairs that are obtained by different distance metrics. In the task of change detection, a suitable distance metric must increase the distance between the changed feature pairs and decrease the distance between the unchanged feature pairs.

\begin{figure}[htb]
  \begin{center}
  \subfigure[]{
    \label{fig:subfig:a} 
    \includegraphics[width=0.5in]{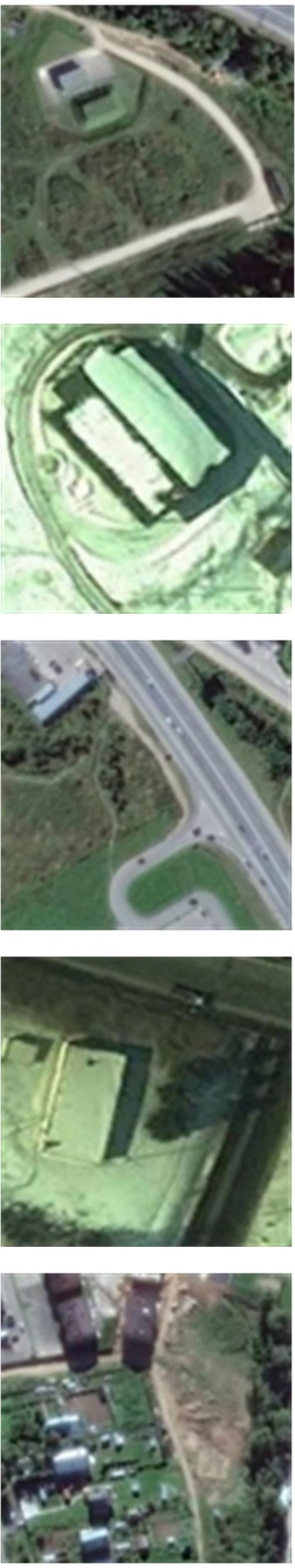}}
  \subfigure[]{
    \label{fig:subfig:b} 
    \includegraphics[width=0.5in]{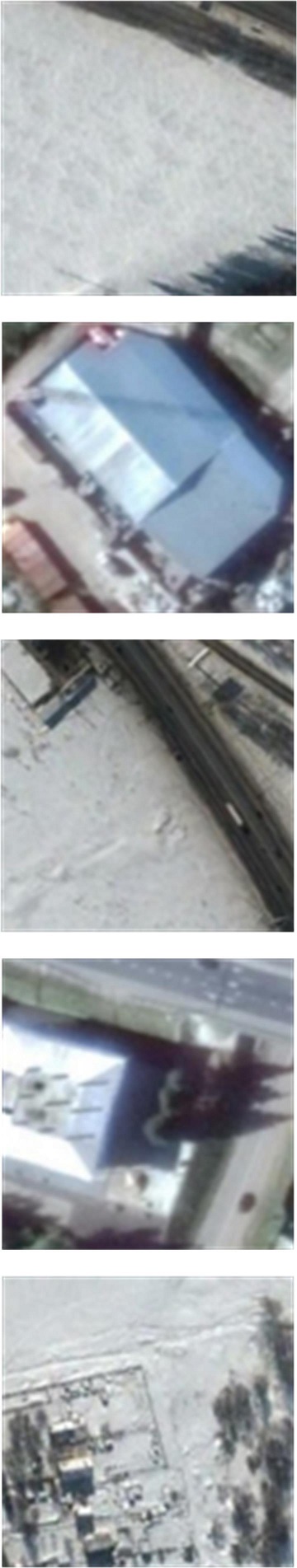}}
  \subfigure[]{
    \label{fig:subfig:c} 
    \includegraphics[width=0.5in]{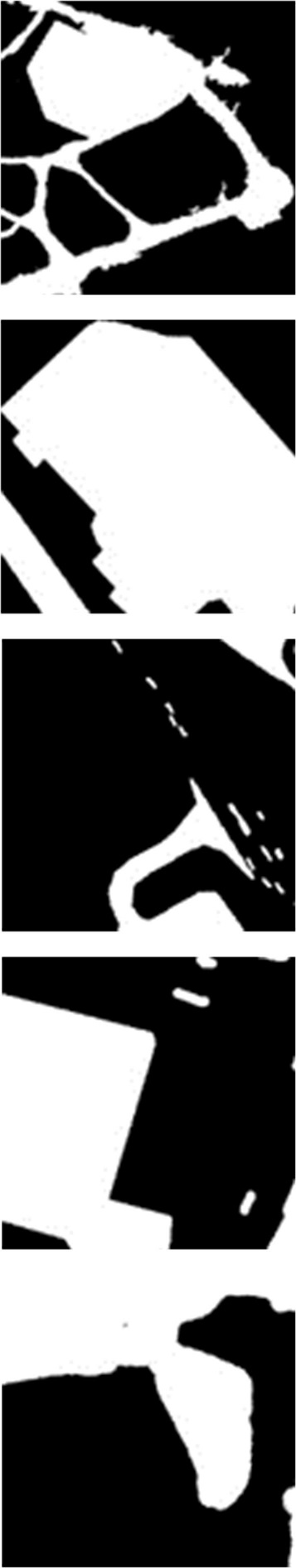}}
  \subfigure[]{
    \label{fig:subfig:d} 
    \includegraphics[width=0.5in]{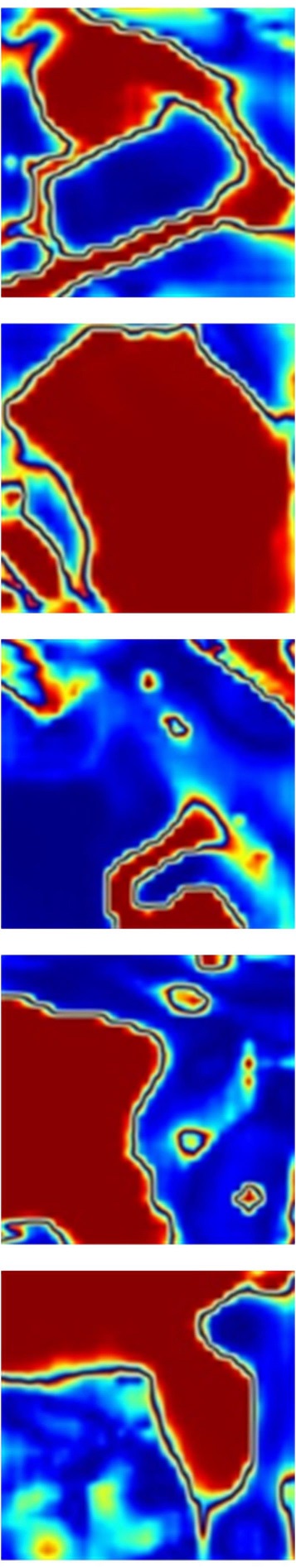}}
  \subfigure[]{
    \label{fig:subfig:e} 
    \includegraphics[width=0.5in]{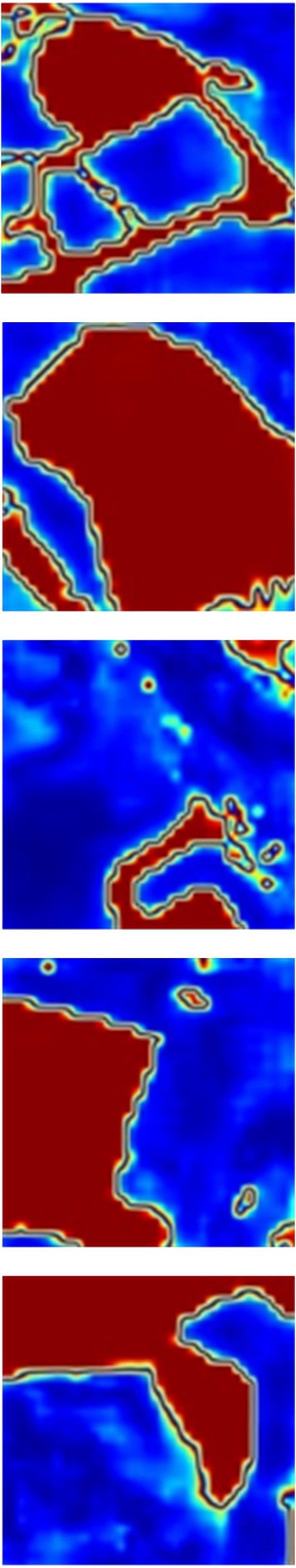}}    
  \caption{Distance map visualizations for various layers: (a) unchanged image, (b) changed image, (c) label, (d) distance map of DASNet (VGG16) with the cosine similarity, (e) distance map of DASNet (VGG16)  with the l2 distance. The changed parts are shown in white, while the unchanged parts are shown in black.}
  \label{heat} 
  \end{center}
\end{figure}

To obtain a suitable distance metric, we evaluated the impacts of the l2 distance and the cosine similarity on the results of the change detection task. We visualize the results from various distance metrics. As shown in Fig.~\ref{heat}, we visualized the output of two trained DASNet networks, which were all set up in the same way except for different distance metrics, and we obtained the distance maps based on various distance metrics. For the same feature layer, the background of the receiving map that was obtained using the cosine similarity was noisier, and its performance in describing the changing boundary was weaker. Thus, the performance of this metric in excluding noisy changes was weak. The resulting graph that was obtained by using the second-order Euclidean distance had a cleaner background and a higher contrast in the front background, which endowed the model with stronger change recognition performance.

In the change detection task, there were far more unchanging sample pairs than changing sample pairs. The traditional contrastive loss has an unbalanced penalty for unchanged feature pairs and changed feature pairs. As a result, the network with the traditional contrastive loss performed weakly in discerning change information. We propose the WDMC loss to strengthen the network's performance in the identification of change information.

\begin{figure}[htb]
  \begin{center}
  \includegraphics[width=2.8in]{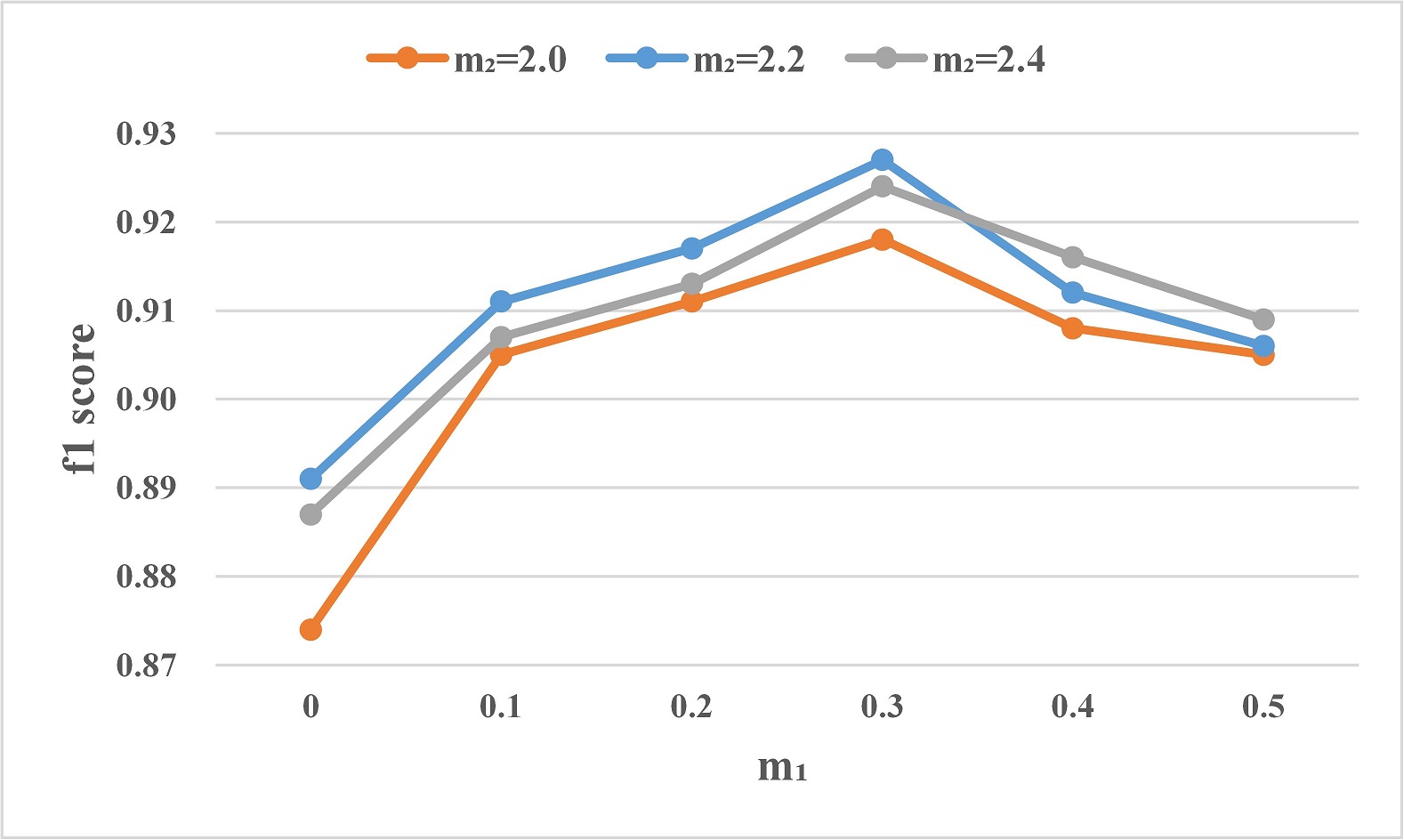}
  \caption{The influence of the values of  $m_{1}$ and $m_{2}$ on the performance of DSANet.}\label{pm}
  \end{center}
\end{figure}

\begin{table}
\begin{center}
\caption{Influence of different values of $\lambda_{1}$, $\lambda_{2}$ and $\lambda_{3}$.}
\label{tabsd}
\begin{tabular}{lllllll}
\hline
\multicolumn{1}{c}{$\lambda_{1}$ } & $\lambda_{2}$     & $\lambda_{3}$    & Rec   & Pre   & F1    & OA    \\ \hline
0                     & 0    & 1   & 0.924 & 0.921 & 0.922 & 0.981 \\
0.25                  & 0.25 & 0.5 & 0.927 & 0.925 & 0.926 & 0.982 \\
1                     & 1    & 1   & 0.932 & 0.922 & 0.927 & 0.982 \\ \hline
\end{tabular}
\end{center}
\end{table}

Since the selection of the values of $m_{1}$ and $m_{2}$ will have a great influence on the results of the network, many experiments were designed to find the values of $m_{1}$ and $m_{2}$ with the best performance. Fig.~\ref{pm} shows the performance of a DSANet network using ResNet50 as the backbone on the CDD dataset with different m1 and m2 values. We also tested the effect of different values of $\lambda_{1}$, $\lambda_{2}$ and $\lambda_{3}$ on the CDD data set to the final result. As seen in Table \ref{tabsd}, the appropriate choice of $\lambda_{1}$, $\lambda_{2}$ and $\lambda_{3}$ improved the accuracy of the model to a certain extent. Table \ref{tab1} shows the comparison between the improved loss function and the traditional contrastive loss, and precision, recall, F1 score, and OA were used as judgment indicators to evaluate the performance.

\begin{table}
\begin{center}
\caption{Loss comparison studies on the CDD dataset.}
\label{tab1}
\setlength{\tabcolsep}{1.0mm}{
\begin{tabular}{llllll}
\hline
\multicolumn{1}{c}{\textbf{Method}} & \multicolumn{1}{c}{\textbf{BaseNet}} & \multicolumn{1}{c}{\textbf{Rec}}   & \multicolumn{1}{c}{\textbf{Pre}}   & \multicolumn{1}{c}{\textbf{F1}}    & \multicolumn{1}{c}{\textbf{OA}}    \\ \hline
Siam-Conv with contrastive loss     & VGG16                                & \multicolumn{1}{c}{0.822}          & \multicolumn{1}{c}{0.913}          & \multicolumn{1}{c}{0.859}          & \multicolumn{1}{c}{0.967}          \\
Siam-Conv with WDMC loss(cos)       & VGG16                                & \multicolumn{1}{c}{0.889}          & \multicolumn{1}{c}{0.827}          & \multicolumn{1}{c}{0.856}          & \multicolumn{1}{c}{0.963}          \\
Siam-Conv with WDMC loss(l2)        & VGG16                                & \multicolumn{1}{c}{0.896}          & \multicolumn{1}{c}{0.888}          & \multicolumn{1}{c}{0.892}          & \multicolumn{1}{c}{0.973}          \\
DASNet with contrastive loss        & VGG16                                & 0.844                              & 0.915                              & 0.878                              & 0.971                              \\
DASNet with WDMC loss(cos)          & VGG16                                & 0.896                              & 0.841                              & 0.871                              & 0.970                              \\
DASNet with WDMC loss(l2)           & VGG16                                & \multicolumn{1}{c}{\textbf{0.925}} & \multicolumn{1}{c}{\textbf{0.914}} & \multicolumn{1}{c}{\textbf{0.919}} & \multicolumn{1}{c}{\textbf{0.980}} \\ \hline
Siam-Conv with contrastive loss     & ResNet50                             & 0.841                              & 0.915                              & 0.876                              & 0.971                              \\
Siam-Conv with WDMC loss(cos)       & ResNet50                             & 0.893                              & 0.841                              & 0.866                              & 0.969                              \\
Siam-Conv with WDMC loss(l2)        & ResNet50                             & 0.902                              & 0.898                              & 0.900                              & 0.975                              \\
DSANet with contrastive loss        & ResNet50                             & 0.878                              & 0.905                              & 0.891                              & 0.973                              \\
DASNet with WDMC loss(cos)          & ResNet50                             & 0.901                              & 0.879                              & 0.890                              & 0.973                              \\
DASNet with WDMC loss(l2)           & ResNet50                             & \textbf{0.932}                     & \textbf{0.922}                     & \textbf{0.927}                     & \textbf{0.982}                     \\ \hline
\end{tabular}}
\end{center}
\end{table}

According to the results in Table \ref{tab1}, when we used VGG16 as the backbone, compared with the traditional contrastive loss, the recall and F1 score of our proposed WDMC loss function with l2 distance improved by 7.4\% and 3.3\%, respectively, and OA improved by 0.6\%, while other experimental conditions remained unchanged. When we used ResNet50 as the backbone, for Siam-Conv, compared with the traditional contrastive loss, the recall and F1 score of our proposed WDMC loss function with l2 distance improved by 6.1\% and 2.4\%, respectively, and OA improved by 0.4\%, while other experimental conditions remained unchanged. We also carried out experiments on the improved loss and traditional contrastive loss based on the proposed method, and they also proved the effectiveness of the proposed WDMC loss. Therefore, the use of the proposed WDMC loss can improve the performance of the network.

\begin{figure}[htb]
  \begin{center}
  \subfigure[]{
    \label{fig:subfig:al1} 
    \includegraphics[width=0.5in]{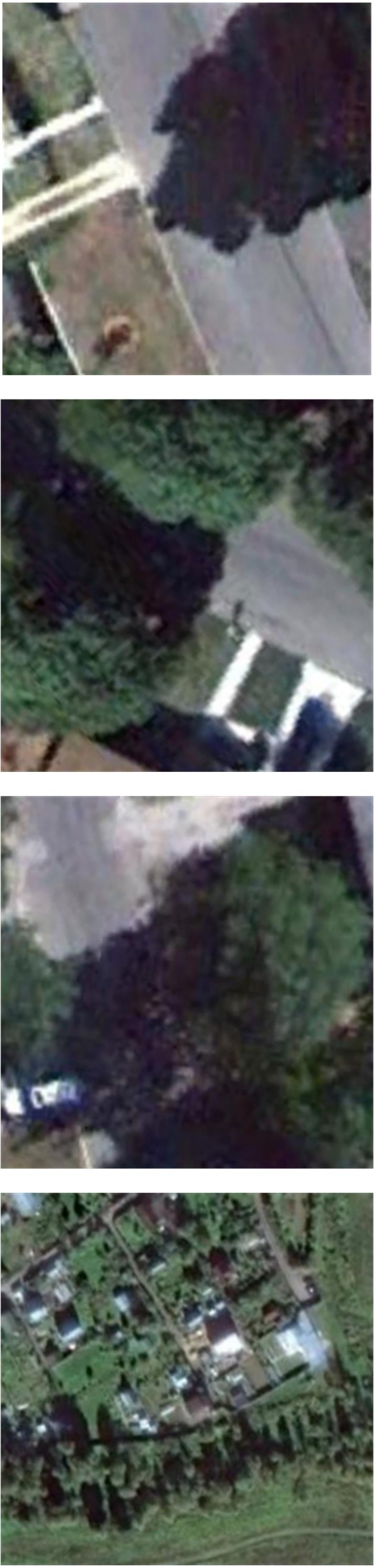}}
  \subfigure[]{
    \label{fig:subfig:bl1} 
    \includegraphics[width=0.5in]{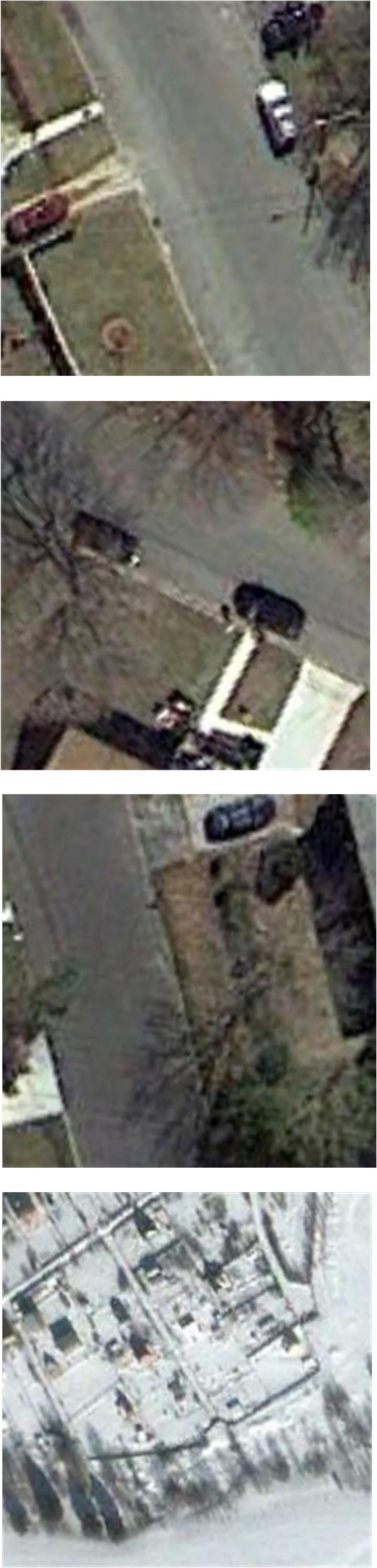}}
  \subfigure[]{
    \label{fig:subfig:cl1} 
    \includegraphics[width=0.5in]{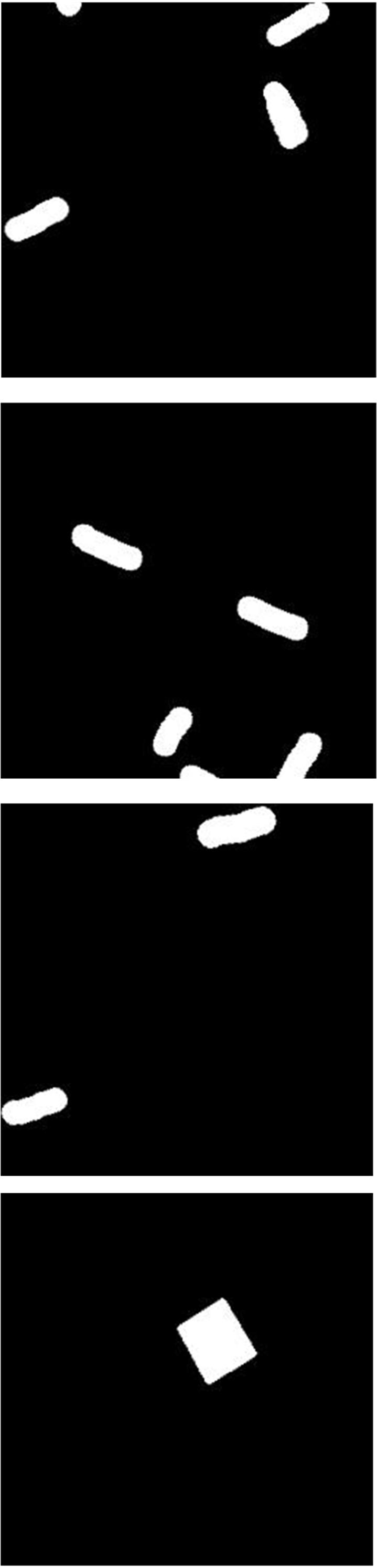}}
  \subfigure[]{
    \label{fig:subfig:dl1} 
    \includegraphics[width=0.5in]{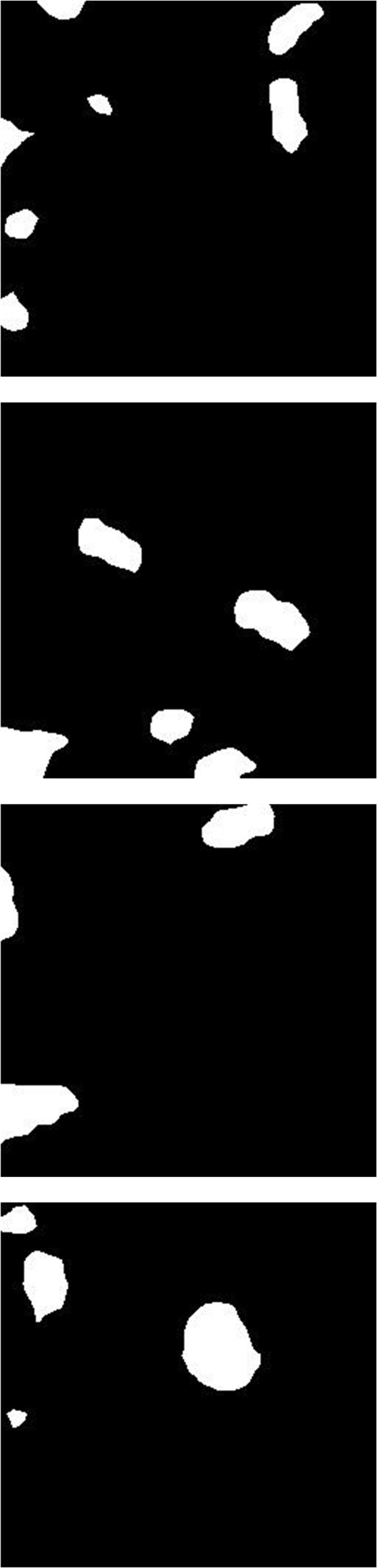}}
  \subfigure[]{
    \label{fig:subfig:el1} 
    \includegraphics[width=0.5in]{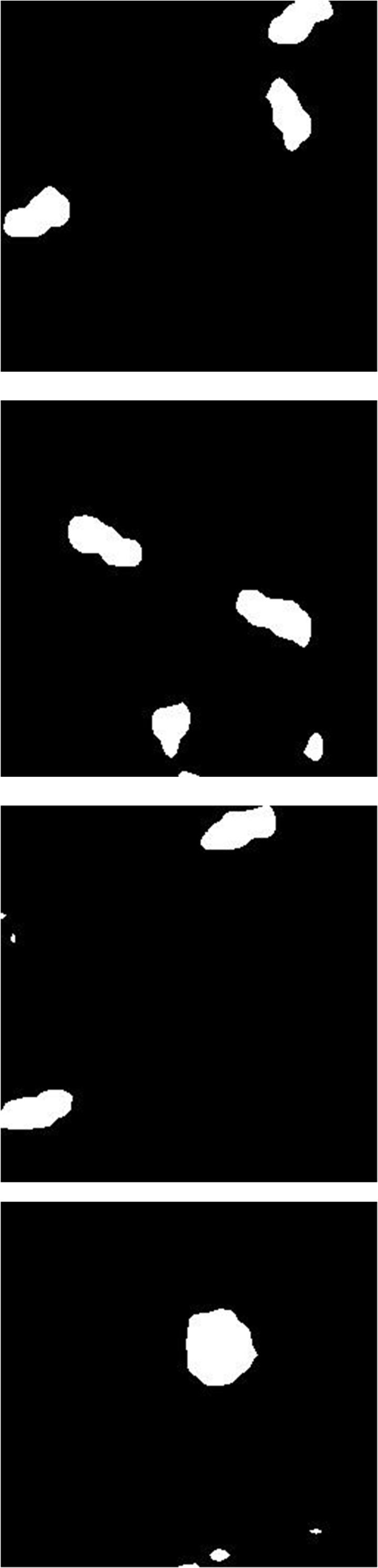}}
  \caption{Visualization results of Siam-Conv (ResNet50) using different losses: (a) unchanged image, (b) changed image, (c) label, (d) result of Siam-Conv (ResNet50) with contrastive loss, (e) result of Siam-Conv (ResNet50) with WDMC loss. The changed parts are shown in white, while the unchanged parts are shown in black.}
  \label{wls} 
  \end{center}
\end{figure}

To verify the robustness of the proposed loss function to pseudo-changes, we visualized the output results of Siam-Conv using traditional contrastive loss. Similarly, we also visualized the output results of Siam-Conv using WDMC loss. The use of Siam-Conv is to avoid the influence of the dual attention mechanism on the results. It can be clearly seen in Fig.~\ref{wls} that the network trained with WDMC loss was more resistant to pseudo-changes. The network trained with WDMC loss was better able to distinguish the pseudo-changes due to seasonal changes, sensor changes, etc. However, the traditional contrastive loss could not do this well.

\subsection{Ablation Study for Attention Modules}
\begin{figure}[H]
  \begin{center}
  \includegraphics[width=2.8in]{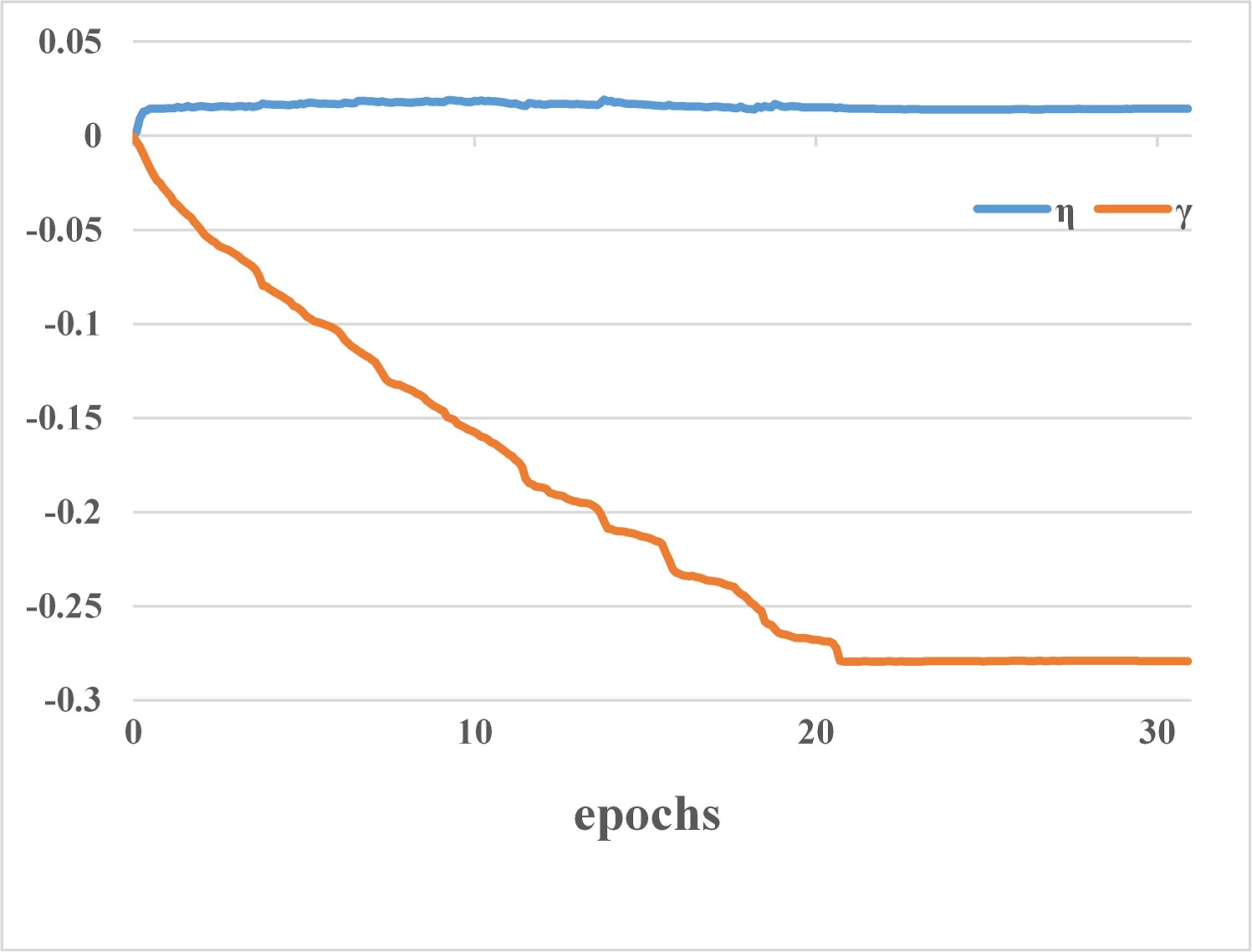}
  \caption{Change process of the two parameters $\eta$  and $\gamma$ in 30 epochs}\label{psb}
  \end{center}
\end{figure}

To better focus on the change regions and to improve the recognition accuracy of the network, on the basis of the WDMC loss, a dual attention module was incorporated for the identification of long-range dependencies to obtain better feature representations. We designed ablation experiments to evaluate the performance of the dual attention mechanism. As Fig.~\ref{psb} shows, we also give the change process of the two parameters $\eta$  and $\gamma$ in 30 epochs.

According to Table \ref{tab2}, the dual attention module improved the model’s performance comprehensively. Compared with the baseline Siamese network (a Siamese network that is based on VGG16 or ResNet50), for using VGG16 as the backbone, the recall, precision, F1 score and OA of the network with the spatial attention module were 0.919, 0.901, 0.910 and 0.978, respectively, which correspond to increases of 2.3\%, 1.3\%, 1.8\%, and 0.5\%, respectively. For using ResNet50 as the backbone, the recall, precision, F1 score, and OA of the network with the spatial attention module were 0.927, 0.903, 0.916, and 0.979, respectively, which correspond to increases of 2.5\%, 0.5\%, 1.6\%, and 0.5\%, respectively. The recall, precision, F1 score, and OA of the network that used VGG16 as the backbone with the channel attention module were 0.922, 0.892, 0.906, and 0.976, respectively, which correspond to increases of 2.6\%, 0.4\%, 1.4\%, and 0.3\%, respectively. The recall, precision, F1 score, and OA of the network that used ResNet50 as the backbone with the channel attention module were 0.930, 0.894, 0.912, and 0.978, respectively, which correspond to increases of 2.8\%,.0.4\%, 1.2\%, and 0.4\%, respectively. After both the spatial attention module and the channel attention module were added into the network, the network that used VGG16 performance was further improved. The recall, precision, F1 score and OA were 0.925, 0.914, 0.919 and 0.980, respectively, which correspond to increases of 2.9\%, 2.6\%, 3.7\% and 0.7\%, respectively. The network that used ResNet50 performance was further improved. The recall, precision, F1 score and OA were 0.932, 0.922, 0.927 and 0.982, respectively, which correspond to increases of 3.0\%, 2.4\%, 2.7\% and 0.6\%, respectively.

\begin{table}
\begin{center}
\caption{Attention mechanism ablation research on the CDD dataset, where CAM denotes channel attention and SAM denotes spatial attention.}
\label{tab2}
\setlength{\tabcolsep}{1.5mm}{
\begin{tabular}{llcccccc}
\hline
\multicolumn{1}{c}{\textbf{Method}} & \multicolumn{1}{c}{\textbf{BaseNet}} & \textbf{SAM}         & \textbf{CAM}         & \textbf{Rec}                       & \textbf{Pre}                       & \textbf{F1}                        & \textbf{OA}                        \\
\hline
Siam-Conv                           & VGG16                                &                      &                      & 0.896                              & 0.888                              & 0.892                              & 0.973                              \\
DASNet                              & VGG16                                &  \checkmark          &                      & 0.919                              & 0.901                              & 0.910                              & 0.978                              \\
DASNet                              & VGG16                                &                      & \checkmark           & 0.922                              & 0.892                              & 0.906                              & 0.976                              \\
DASNet                              & VGG16                                &  \checkmark          &\checkmark            & \textbf{0.925}                     & \textbf{0.914}                     & \textbf{0.919}                     & \textbf{0.980}                     \\
\hline
Siam-Conv                           & ResNet50                             &  &   & {0.902}          & {0.898}          & {0.900}          & {0.974}          \\
DASNet                              & ResNet50                             & \checkmark &   & {0.927}          & {0.903}          & {0.916}          & {0.979}          \\
DASNet                              & ResNet50                             &  &  \checkmark & {0.930}          & {0.894}          & {0.912}          & {0.978}          \\
DASNet                              & ResNet50                             & \checkmark & \checkmark  & {\textbf{0.932}} & {\textbf{0.922}} & {\textbf{0.927}} & {\textbf{0.982}}\\
\hline
\end{tabular}}
\end{center}
\end{table}

The addition of a dual attention module to the Siamese network overcomes the problem that the WDMC loss function cannot improve the precision while improving the recall. Both the spatial attention mechanism and the channel attention mechanism improve the accuracy of the network. When these two attention mechanisms are combined, the spatial information and channel information of the features are fully utilized, and the comprehensive performance of the network is improved.

\subsection{Visualization of the Dual Attention Mechanism Effect}

\begin{figure*}[htb]
  \begin{center}
  \subfigure[]{
    \label{fig:subfig:a1} 
    \includegraphics[width=0.8in]{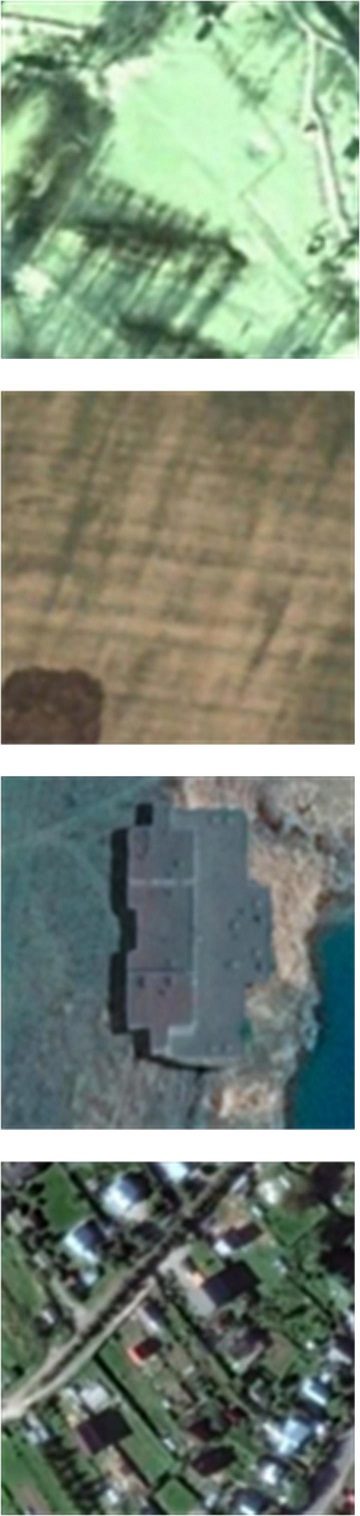}}
  \subfigure[]{
    \label{fig:subfig:b1} 
    \includegraphics[width=0.8in]{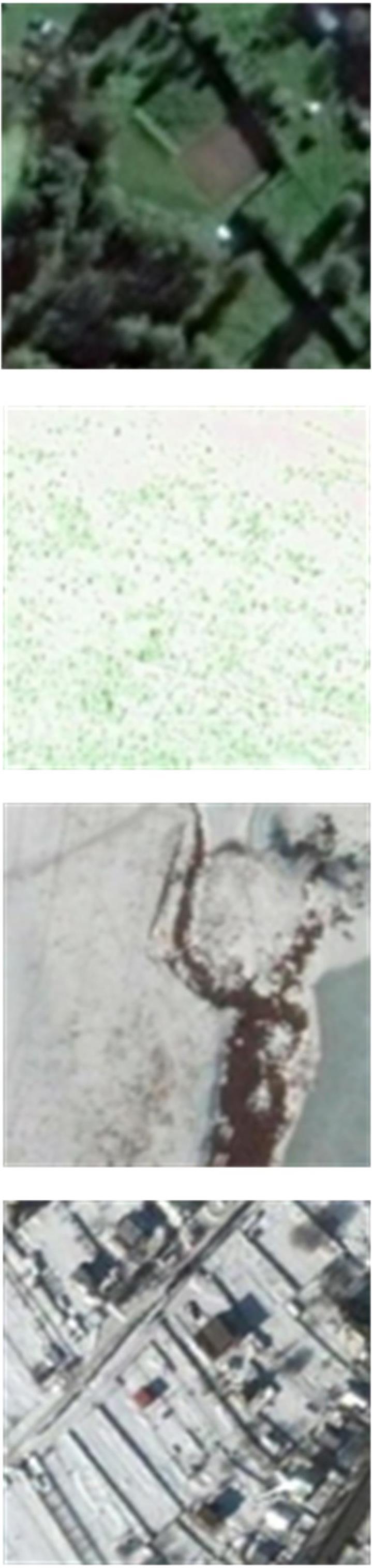}}
  \subfigure[]{
    \label{fig:subfig:c1} 
    \includegraphics[width=0.8in]{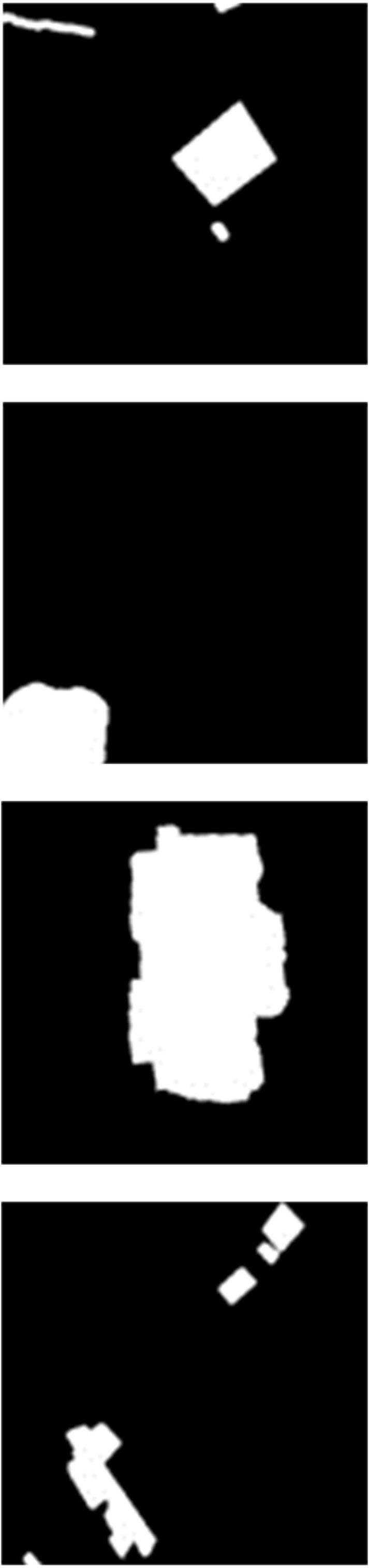}}
  \subfigure[]{
    \label{fig:subfig:d1} 
    \includegraphics[width=0.8in]{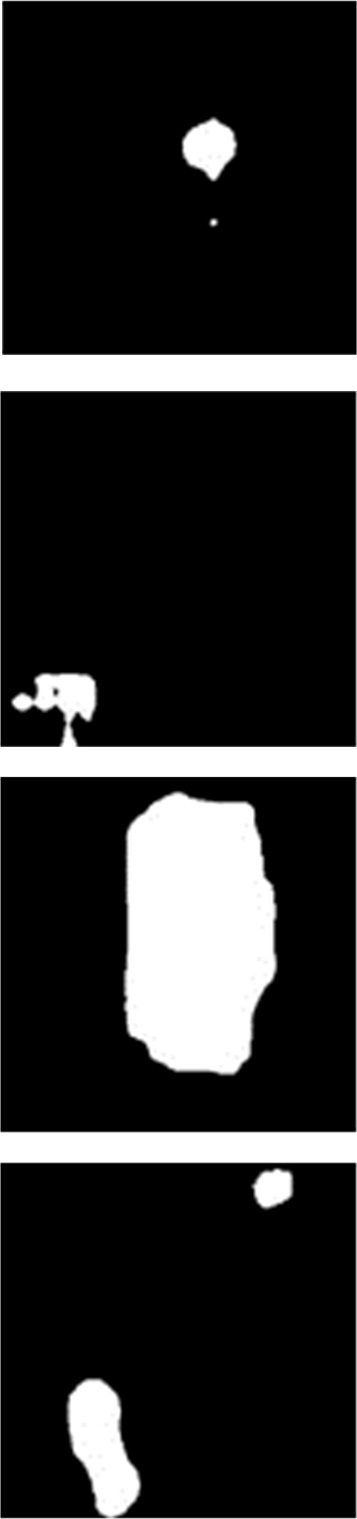}}
  \subfigure[]{
    \label{fig:subfig:e1} 
    \includegraphics[width=0.8in]{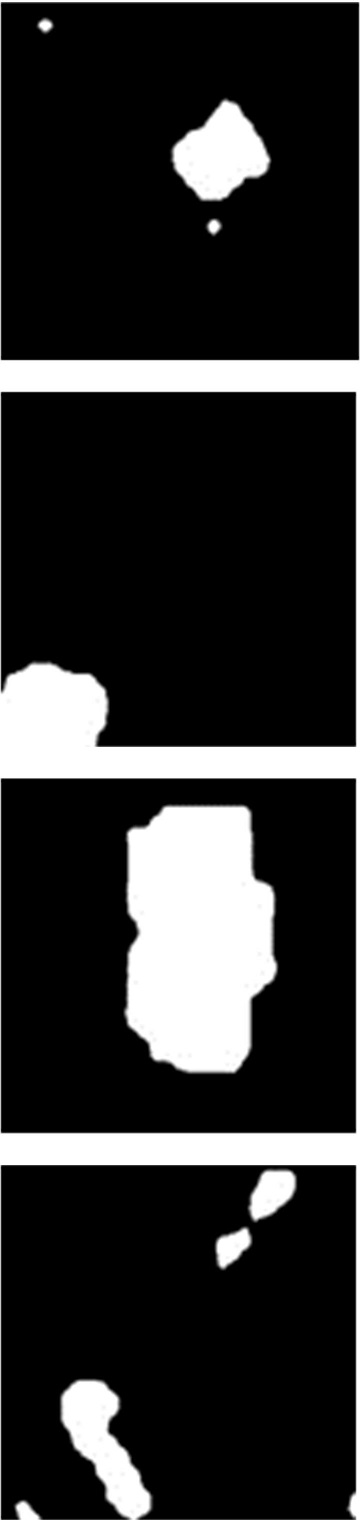}}
  \subfigure[]{
    \label{fig:subfig:F1} 
    \includegraphics[width=0.8in]{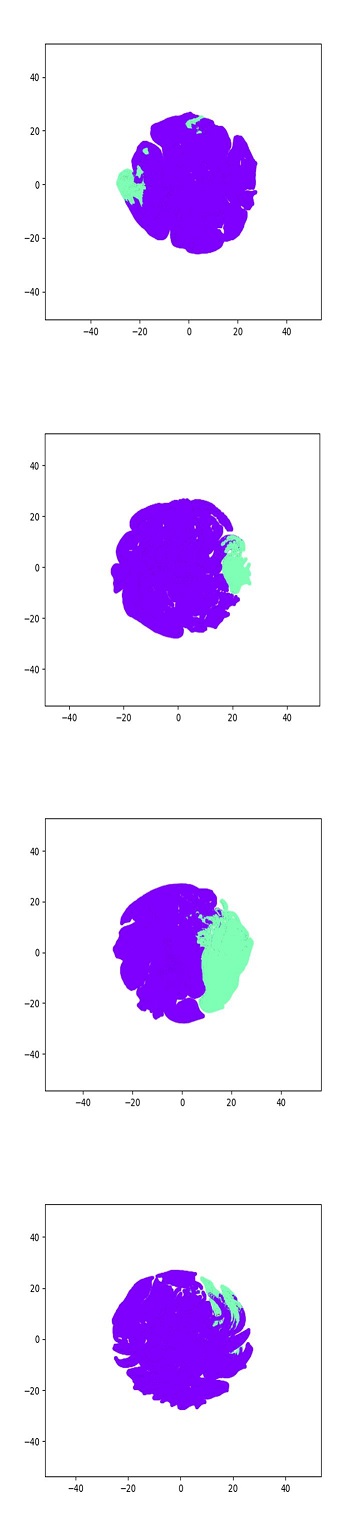}}
  \subfigure[]{
    \label{fig:subfig:g1} 
    \includegraphics[width=0.8in]{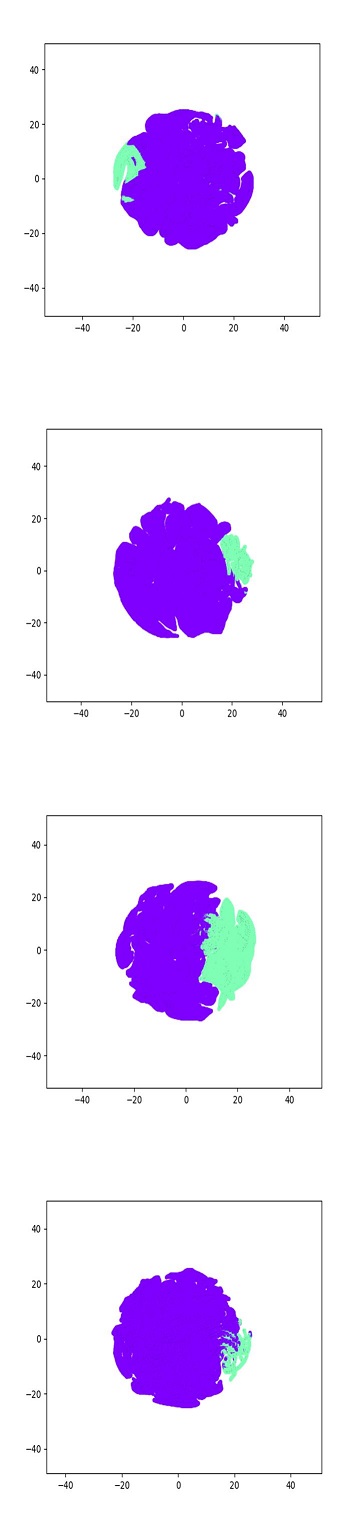}}
  \end{center}
  \caption{ t-SNE visual comparison diagram: (a) unchanged image, (b) changed image, (c) label, (d) result of Siam-Conv, (e) result of DASNet, (f) the last feature layer of Siam-Conv, and (g) the last feature layer of DASNet. The changed parts are shown in white, while the unchanged parts are shown in black.}
  \label{tsne} 
\end{figure*}

To more intuitively reflect the role of the dual attention module, we used the t-SNE \cite{maaten2008visualizing} algorithm to visualize the last feature layer of the Siamese net without adding the dual attention mechanism (Siam-Conv) and the last feature layer of the proposed model (Fig.~\ref{tsne}). In the t-SNE diagram, the purple parts represent the unchanged feature vectors in 2 dimensions, and the green parts represent the changed feature vectors in 2 dimensions. Compared with Siam-Conv, the main advantage of the DASNet model was that changed and unchanged features were both well clustered, with strong discrimination between them. This enlarged the margin between the types of features and effectively increased the accuracy of feature classification.

\begin{figure}[htb]
  \begin{center}
  \subfigure[]{
    \label{fig:subfig:ag1} 
    \includegraphics[width=0.5in]{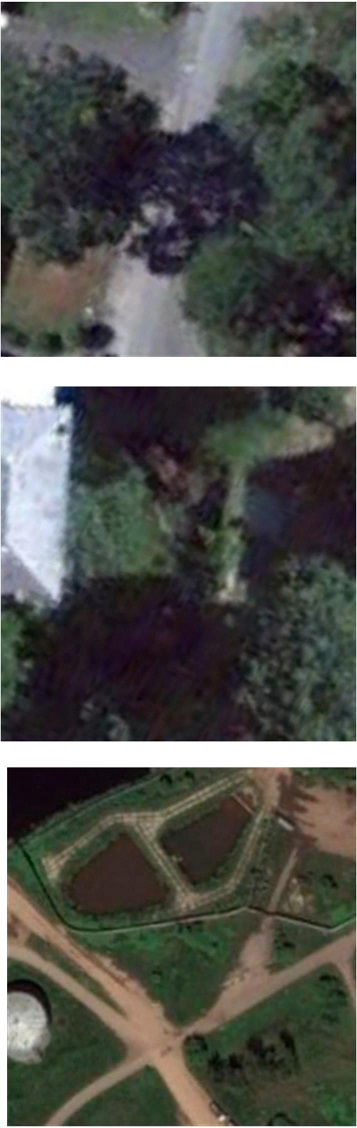}}
  \subfigure[]{
    \label{fig:subfig:bg1} 
    \includegraphics[width=0.5in]{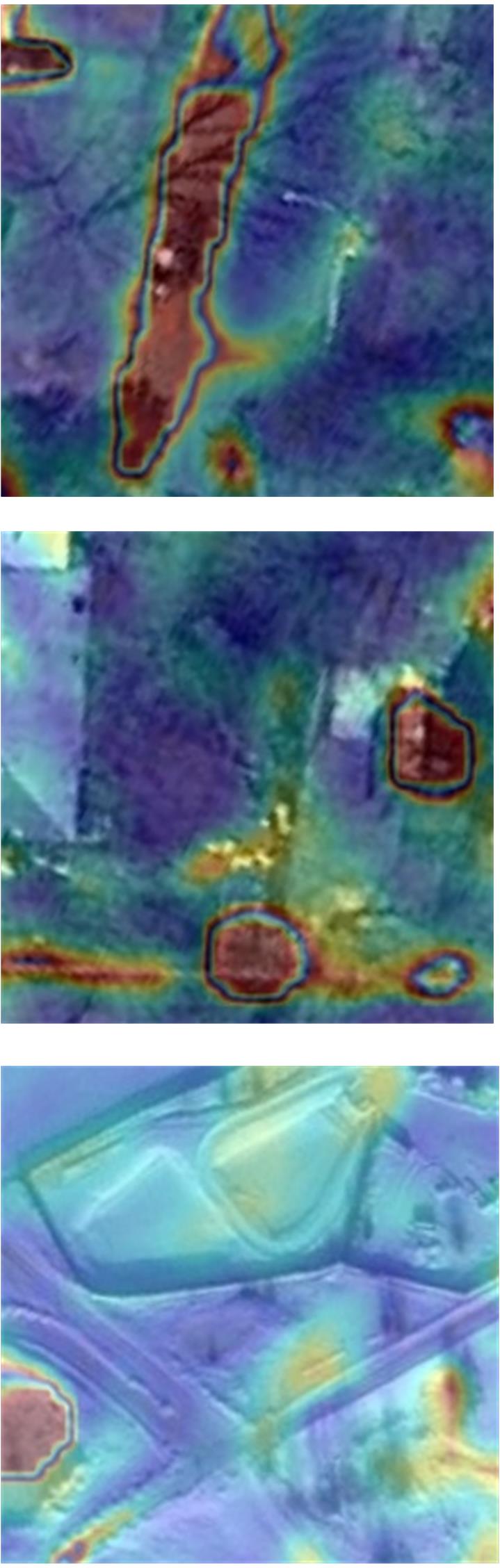}}
  \subfigure[]{
    \label{fig:subfig:cg1} 
    \includegraphics[width=0.5in]{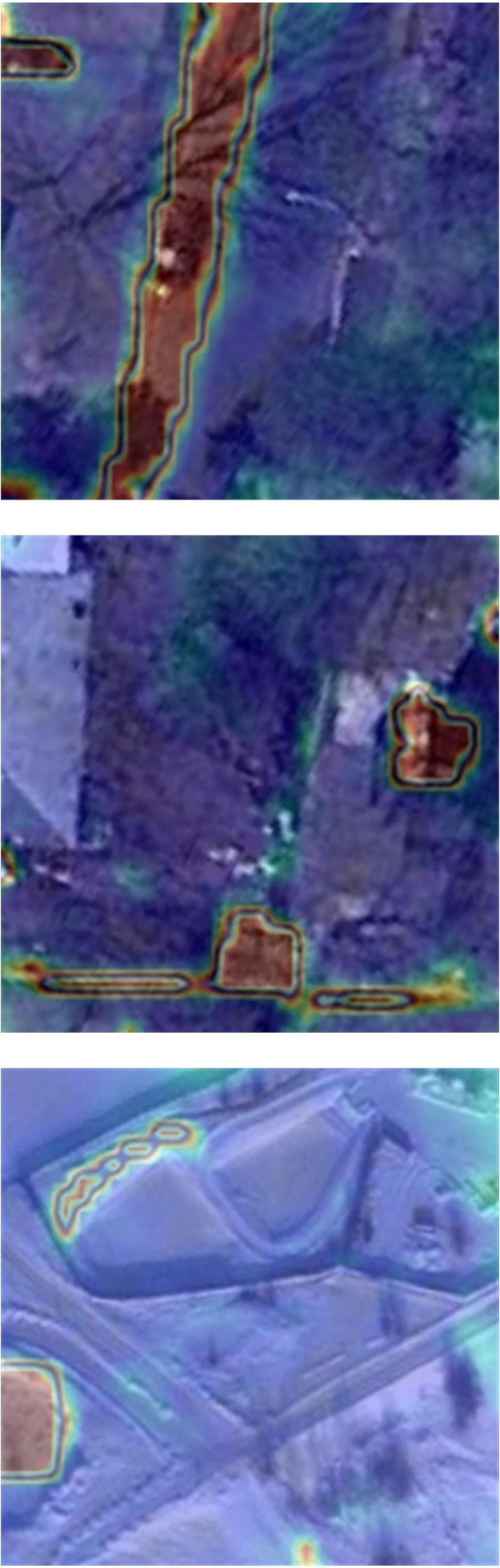}}
  \subfigure[]{
    \label{fig:subfig:dg1} 
    \includegraphics[width=0.5in]{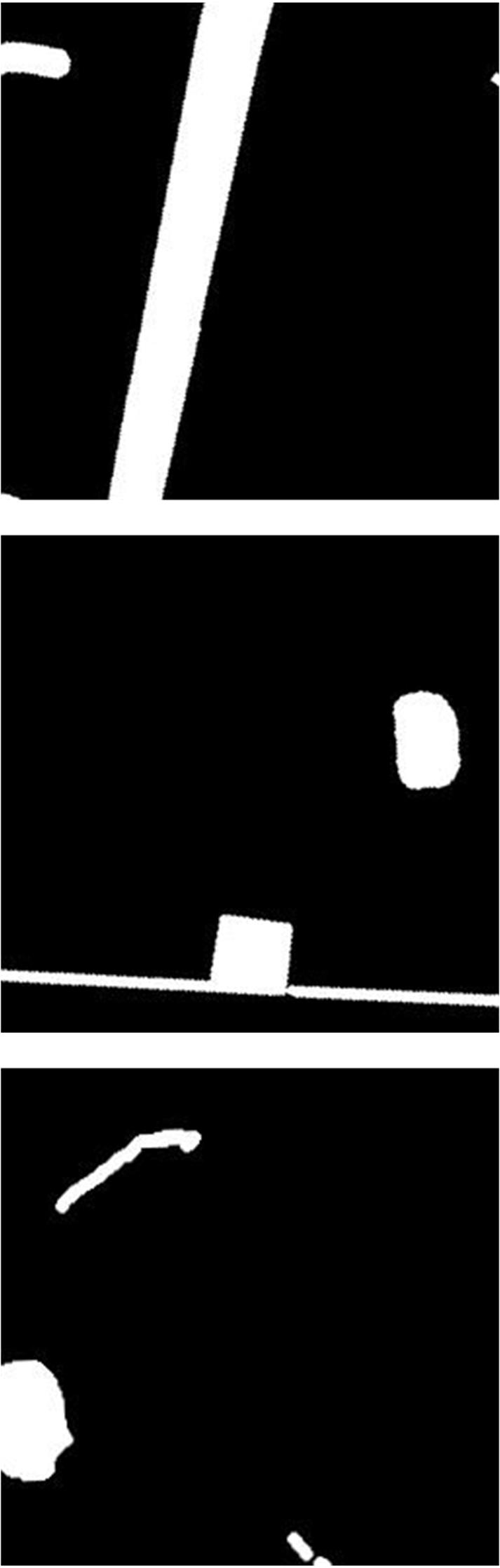}}
  \caption{Grad-CAM visualization results: (a) unchanged image, (b) result of the fusion of the changed image with Grad-CAM mask of the Siam-Conv, (c) result of the fusion of the changed image with Grad-CAM mask of the DASNet, and (d) label. The changed parts are shown in white, while the unchanged parts are shown in black.}
  \label{grad} 
  \end{center}
\end{figure}

To further illustrate the effectiveness of the attention mechanism, we applied the Grad-CAM \cite{2016Grad} to different networks using images from the CDD dataset. Grad-CAM is a visualization method that uses gradients to calculate the importance of spatial positions in the convolutional layer. We compared the visualization results of Siam-Conv and DASNet. Fig.~\ref{grad} illustrates the visualization results. In Fig.~\ref{grad}, we can clearly see that the Grad-CAM mask of DASNet covers the target area better than Siam-Conv. In other words, DASNet can make good use of the information in the target object area and aggregate characteristics from it.

Fig.~\ref{grad} also demonstrates the robustness of the dual attention mechanism to pseudo-changes. Siam-Conv focuses not only on change information but also on pseudo-change due to seasonal changes in trees, agricultural fields, etc. This is a manifestation of the network's lack of resistance to the interference of pseudo-change information. DASNet, however, is more concerned with real changes.

\subsection{Performance Experiment}

To evaluate the performance of the proposed method, we compared it with several image-based deep learning change detection methods: CDnet \cite{alcantarilla2018street} is used in the study of street scene change detection. It is composed of contraction blocks and expansion blocks, and the change map is obtained through a softmax layer. FC-EF \cite{daudt2018fully} stacks image pairs as the input images. It uses the skip connection structure to fuse the shallow and deep features and finally obtains the change map through the softmax layer. FC-Siam-Conc \cite{daudt2018fully} is an extension of FC-EF. Its encoding layers are separated into two streams with equal structure and shared weights. Then, the skip connections are concatenated in the decoder part. FC-Siam-Diff \cite{daudt2018fully} is another extension of FC-EF. It differs from FC-Siam-Conc in that it does not fuse the feature results that are obtained from the encoder streams in the decoder part but calculates the absolute value of the difference between the feature results. BidateNET \cite{papadomanolaki2019detecting} integrates the convolution module in LSTM \cite{hochreiter1997long}into UNet \cite{ronneberger2015u} so that the model can better learn the temporal change pattern.

We conducted comparative experiments with the above methods on the CDD and BCDD datasets.

\begin{table}
\begin{center}
\caption{Results on the CDD dataset.}
\label{tab3}
\begin{tabular}{cllll}
\hline
Method                         & \multicolumn{1}{c}{Rec} & \multicolumn{1}{c}{Pre} & \multicolumn{1}{c}{F1} & \multicolumn{1}{c}{OA} \\ \hline
CDnet                                                       & 0.817          & 0.827          & 0.822          & 0.964          \\
\begin{tabular}[c]{@{}c@{}}FC-EF\end{tabular} & 0.528          & 0.684          & 0.596          & 0.911          \\
\begin{tabular}[c]{@{}c@{}}FC-Siam-Diff\end{tabular} & 0.703         & 0.677          & 0.691          & 0.931          \\
\begin{tabular}[c]{@{}c@{}}FC-Siam-Conc\end{tabular} & 0.666         & 0.710          & 0.687          & 0.925         \\
\begin{tabular}[c]{@{}c@{}}BiDateNet\end{tabular} & 0.894         & 0.901          & 0.898          & 0.975          \\\hline
DSANet(VGG16)                                                     & \textbf{0.925} & \textbf{0.914} & \textbf{0.919} & \textbf{0.980}\\
DSANet(ResNet50)                                                     & \textbf{0.932} & \textbf{0.922} & \textbf{0.927} & \textbf{0.982}
      \\ \hline

\end{tabular}
    
\end{center}
\end{table}

\begin{table}
\begin{center}
    \caption{Results on the BCDD dataset.}
    \label{tab4}
    \begin{tabular}{cllll}
    \hline
    Method                         & \multicolumn{1}{c}{Rec} & \multicolumn{1}{c}{Pre} & \multicolumn{1}{c}{F1} & \multicolumn{1}{c}{OA} \\ \hline
    CDnet                                                       & 0.821          & 0.908          & 0.862          & 0.989          \\
    \begin{tabular}[c]{@{}c@{}}FC-EF\end{tabular} & 0.746          & 0.841          & 0.791          & 0.981          \\
    \begin{tabular}[c]{@{}c@{}}FC-Siam-Diff\end{tabular} & 0.710         & 0.700          & 0.704          & 0.971          \\
    \begin{tabular}[c]{@{}c@{}}FC-Siam-Conc\end{tabular} & 0.736         & 0.631          & 0.679          & 0.966         \\
    \begin{tabular}[c]{@{}c@{}}BiDateNet\end{tabular} & 0.819         & 0.889          & 0.852          & 0.986          \\\hline
    DASNet(VGG16)                                                     & \textbf{0.905} & \textbf{0.892} & \textbf{0.898} & \textbf{0.990}\\
        DASNet(ResNet50)                                                     & \textbf{0.905} & \textbf{0.900} & \textbf{0.910} & \textbf{0.991}
          \\ \hline
    \end{tabular}
\end{center}
\end{table}

\begin{figure*}[htb]
  \begin{center}

  \subfigure[]{
    \label{fig:subfig:a11} 
    \includegraphics[width=0.5in]{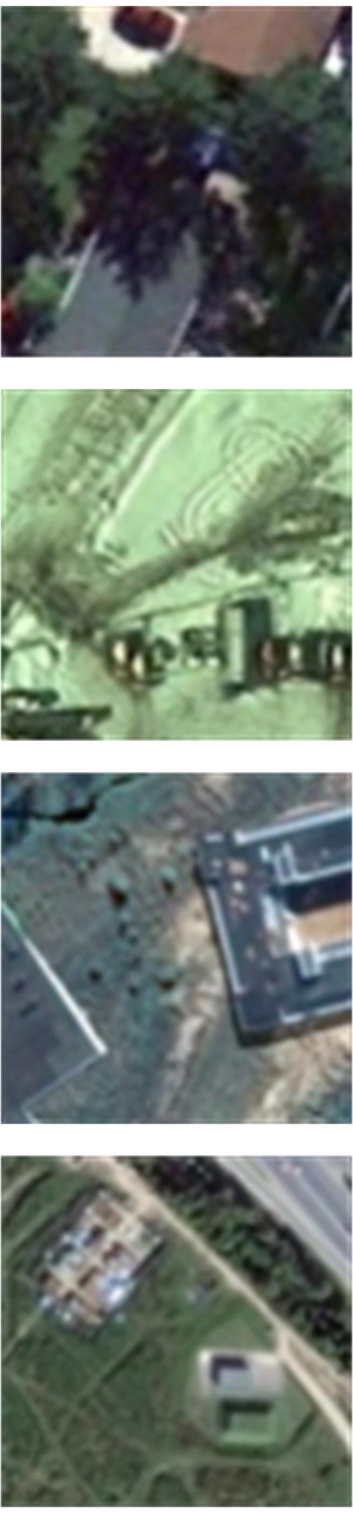}}
  \subfigure[]{
    \label{fig:subfig:b11} 
    \includegraphics[width=0.5in]{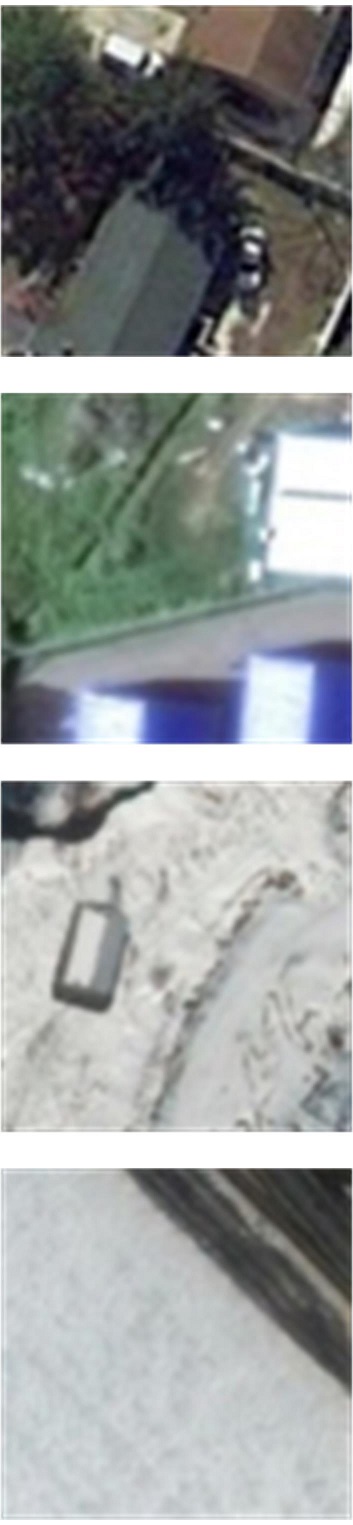}}
  \subfigure[]{
    \label{fig:subfig:c11} 
    \includegraphics[width=0.5in]{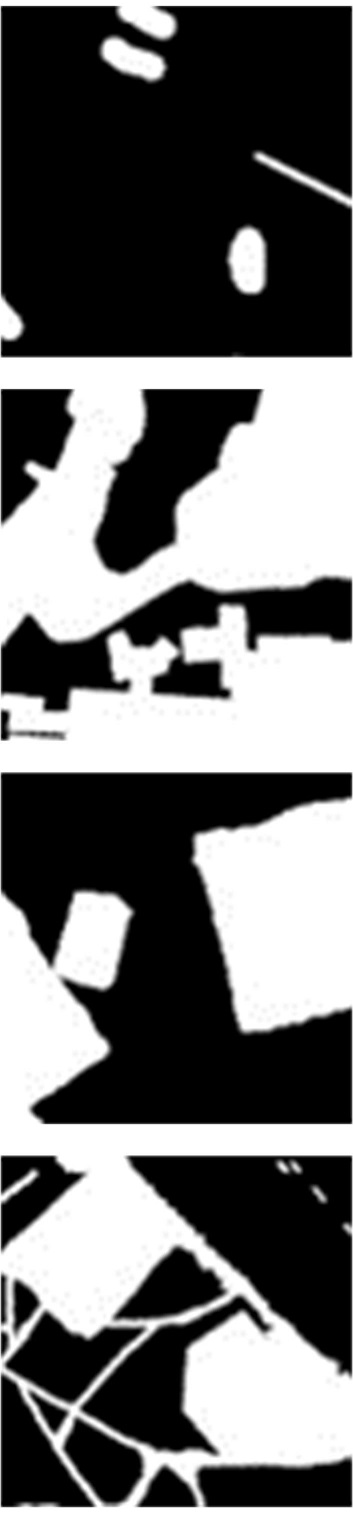}}
  \subfigure[]{
    \label{fig:subfig:d11} 
    \includegraphics[width=0.5in]{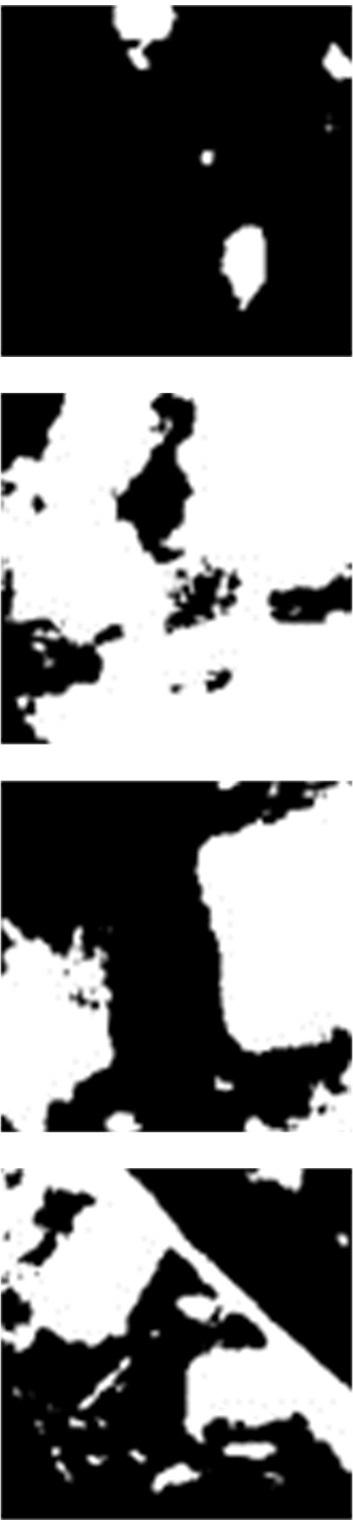}}
  \subfigure[]{
    \label{fig:subfig:e11} 
    \includegraphics[width=0.5in]{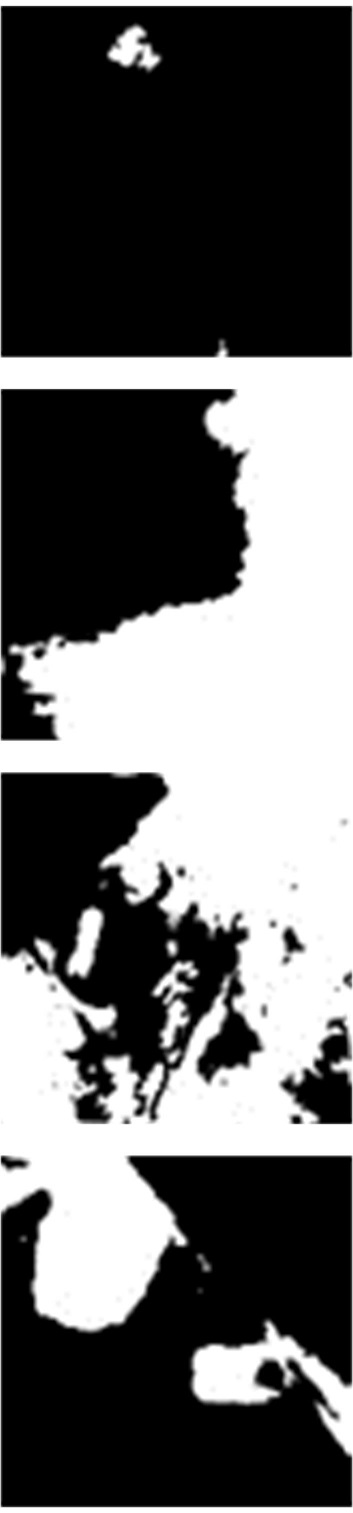}}
  \subfigure[]{
    \label{fig:subfig:F11} 
    \includegraphics[width=0.5in]{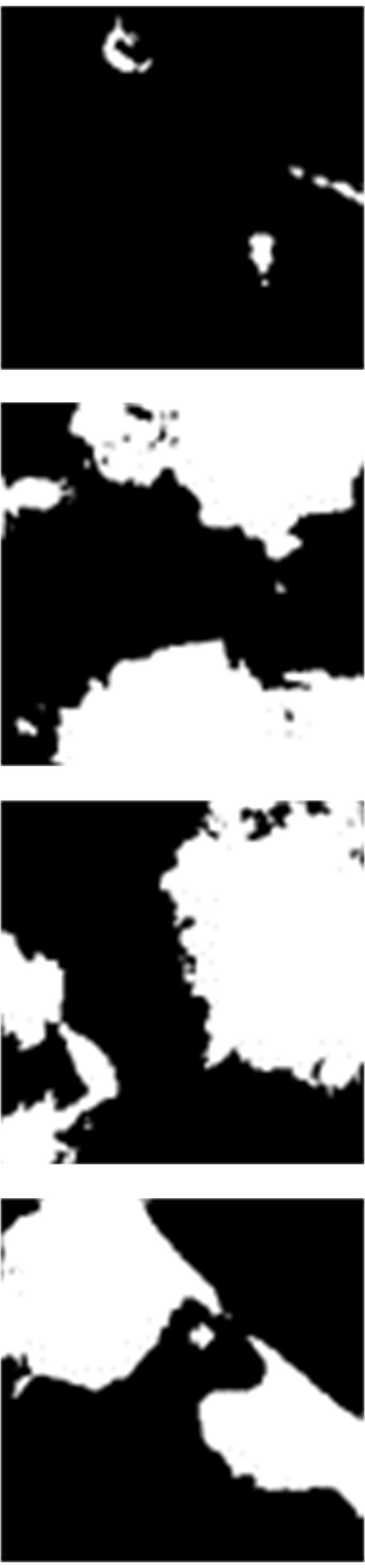}}
  \subfigure[]{
    \label{fig:subfig:g11} 
    \includegraphics[width=0.5in]{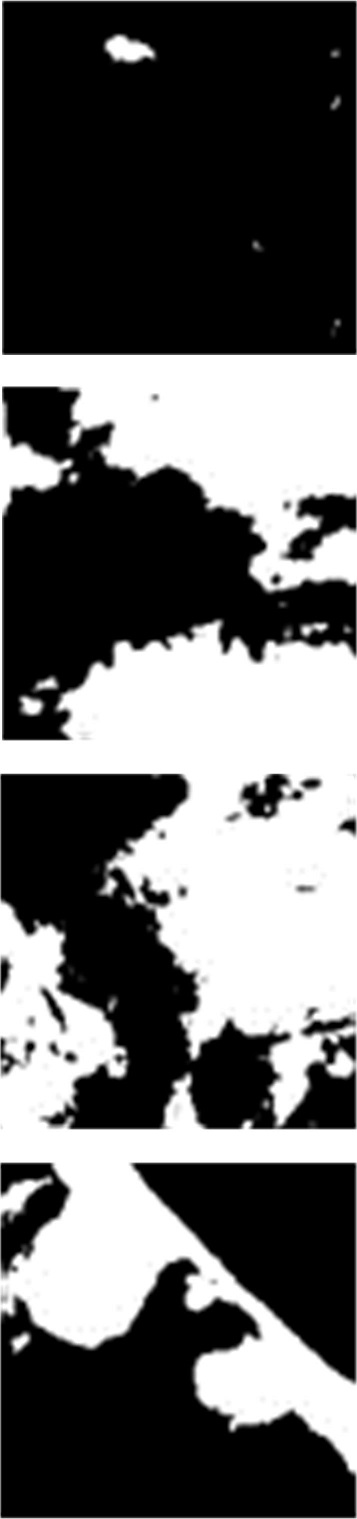}}
  \subfigure[]{
    \label{fig:subfig:h11} 
    \includegraphics[width=0.5in]{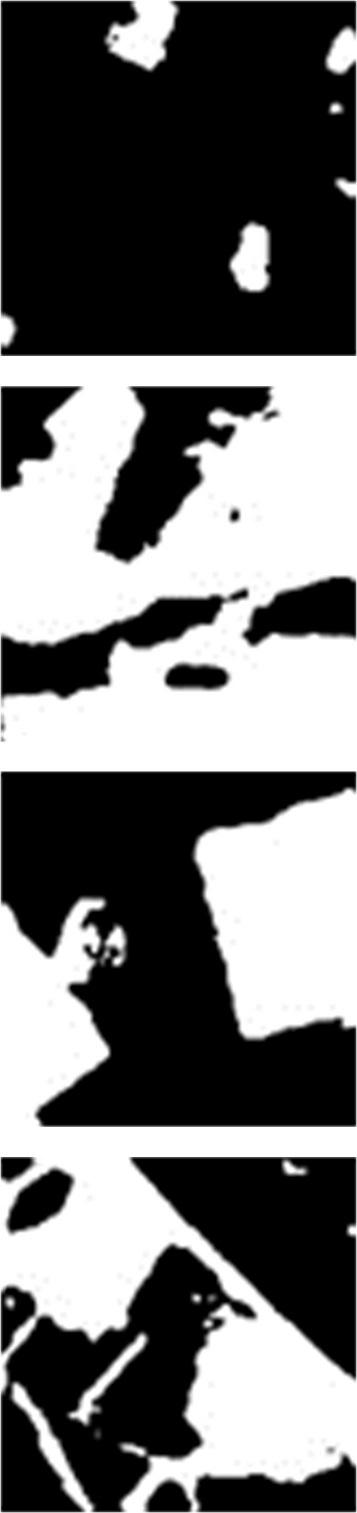}}
  \subfigure[]{
    \label{fig:subfig:i11} 
    \includegraphics[width=0.5in]{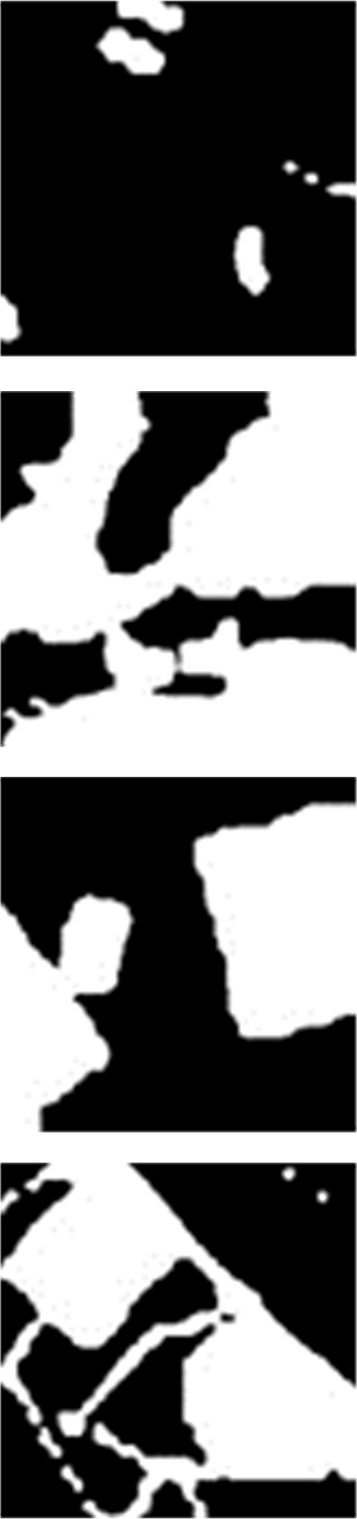}}
   \subfigure[]{
    \label{fig:subfig:j11} 
    \includegraphics[width=0.5in]{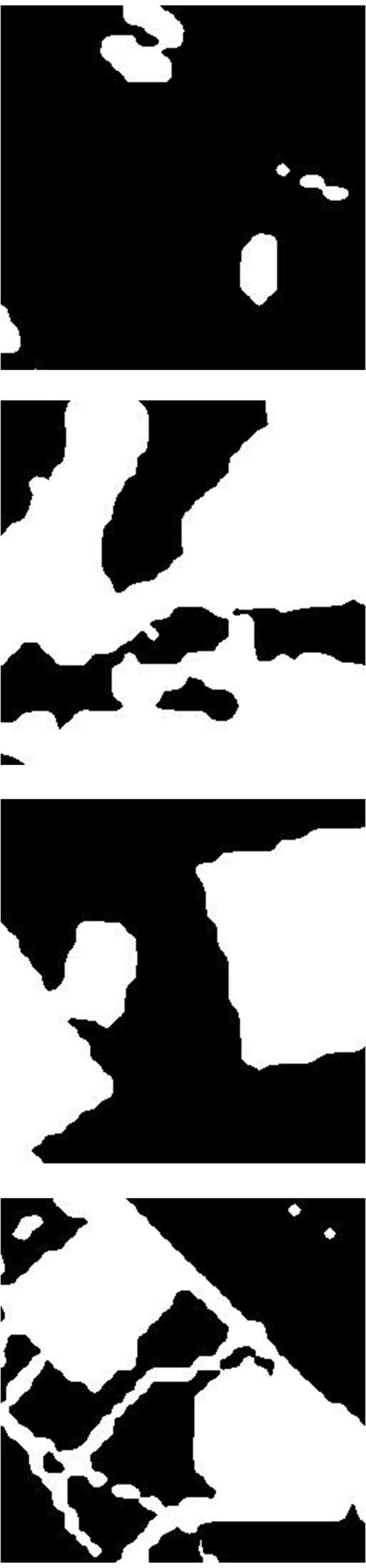}}
  \caption{Visualized comparison of the results of various change detection methods on the CDD dataset: (a) the unchanged image, (b) the changed image, (c) the label, (d) CDnet, (e) FC-EF, (f) FC-Siam-Diff, (g) FC-Siam-Con, (h) BiDateNet, (i) DASNet(VGG16), and (j) DASNet(ResNet50). The changed parts are shown in white, while the unchanged parts are shown in black.}
  \label{CDDr} 

\end{center}
\end{figure*}

\begin{figure*}[htb]
  \begin{center}

  \subfigure[]{
    \label{fig:subfig:a111} 
    \includegraphics[width=0.5in]{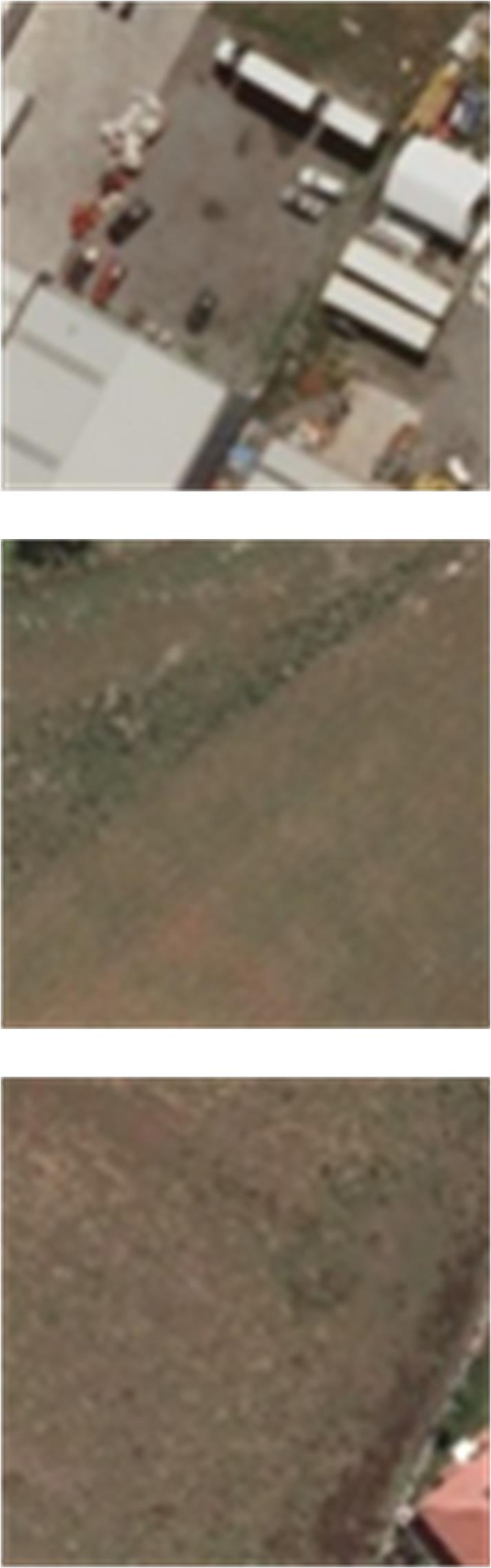}}
  \subfigure[]{
    \label{fig:subfig:b111} 
    \includegraphics[width=0.5in]{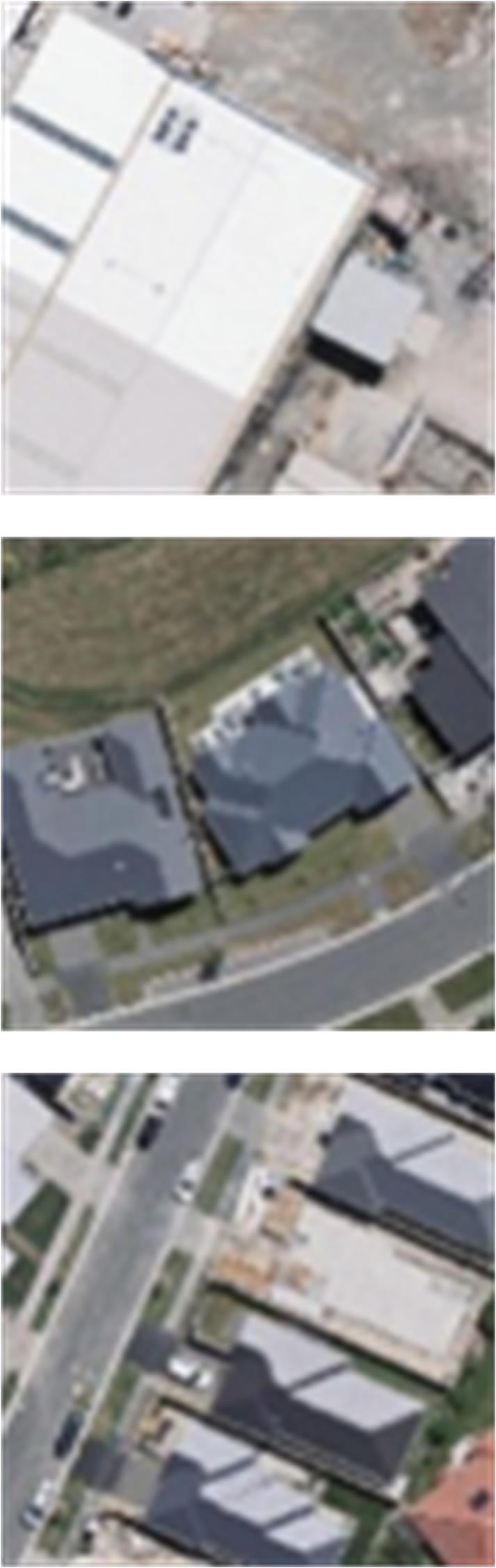}}
  \subfigure[]{
    \label{fig:subfig:c111} 
    \includegraphics[width=0.5in]{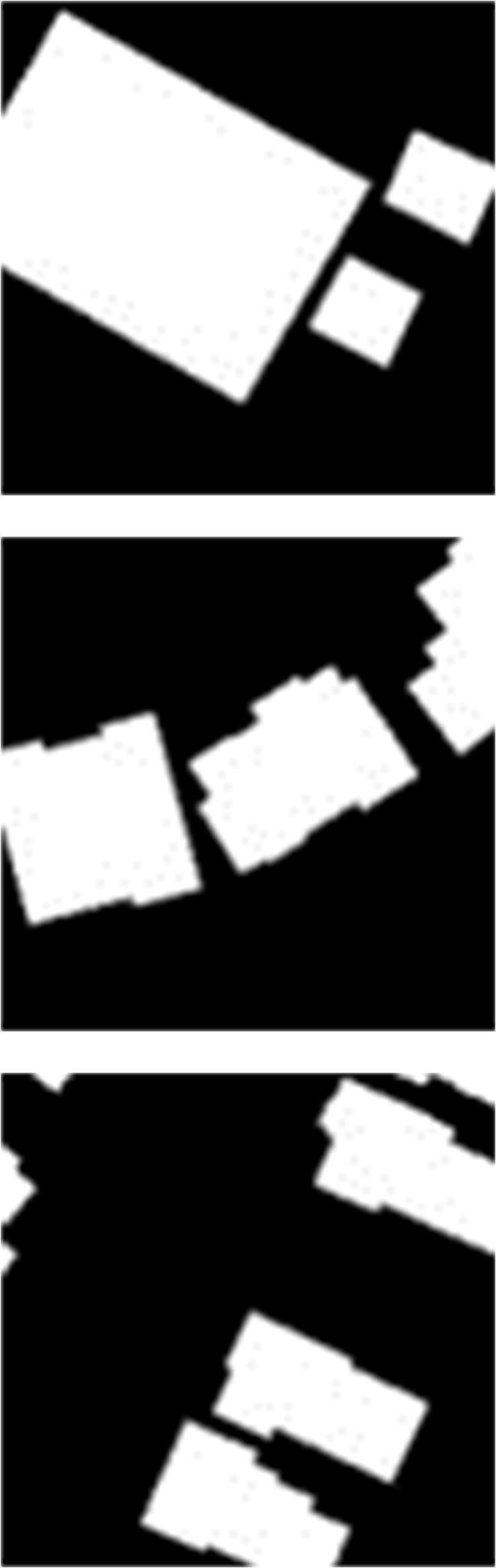}}
  \subfigure[]{
    \label{fig:subfig:d111} 
    \includegraphics[width=0.5in]{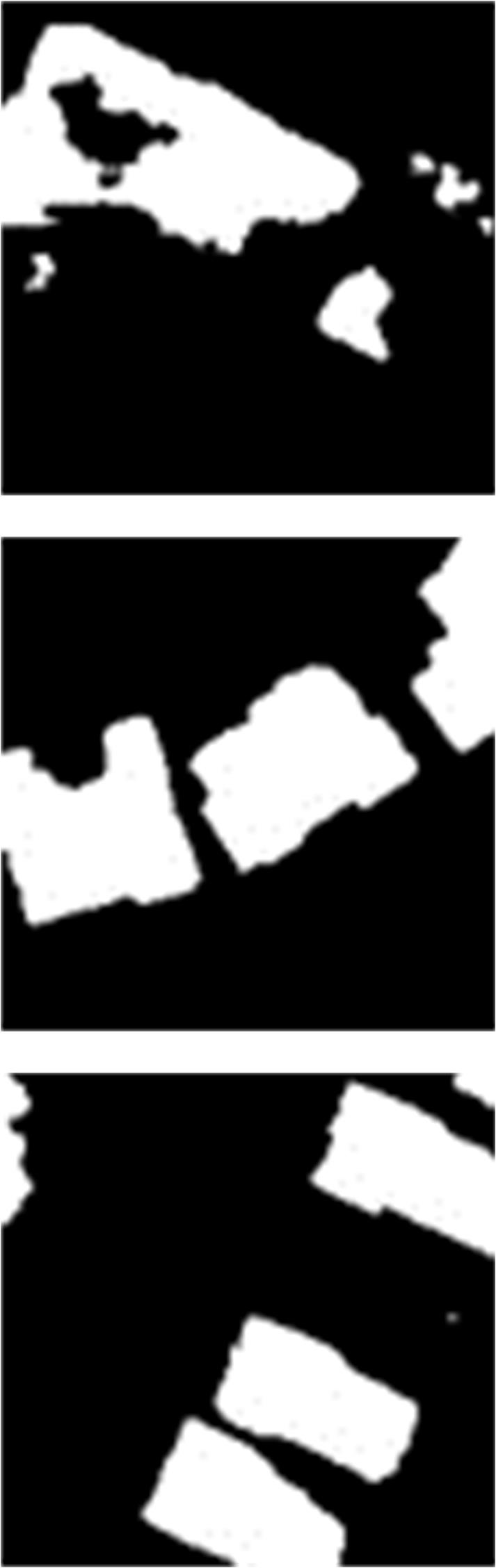}}
  \subfigure[]{
    \label{fig:subfig:e111} 
    \includegraphics[width=0.5in]{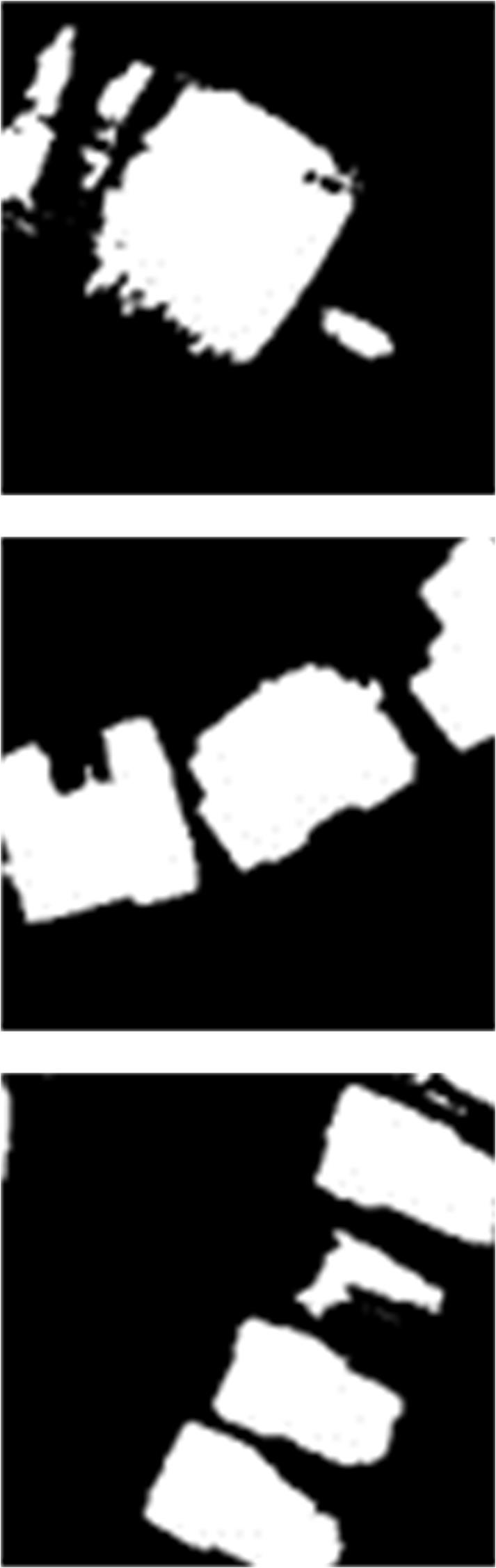}}
  \subfigure[]{
    \label{fig:subfig:F111} 
    \includegraphics[width=0.5in]{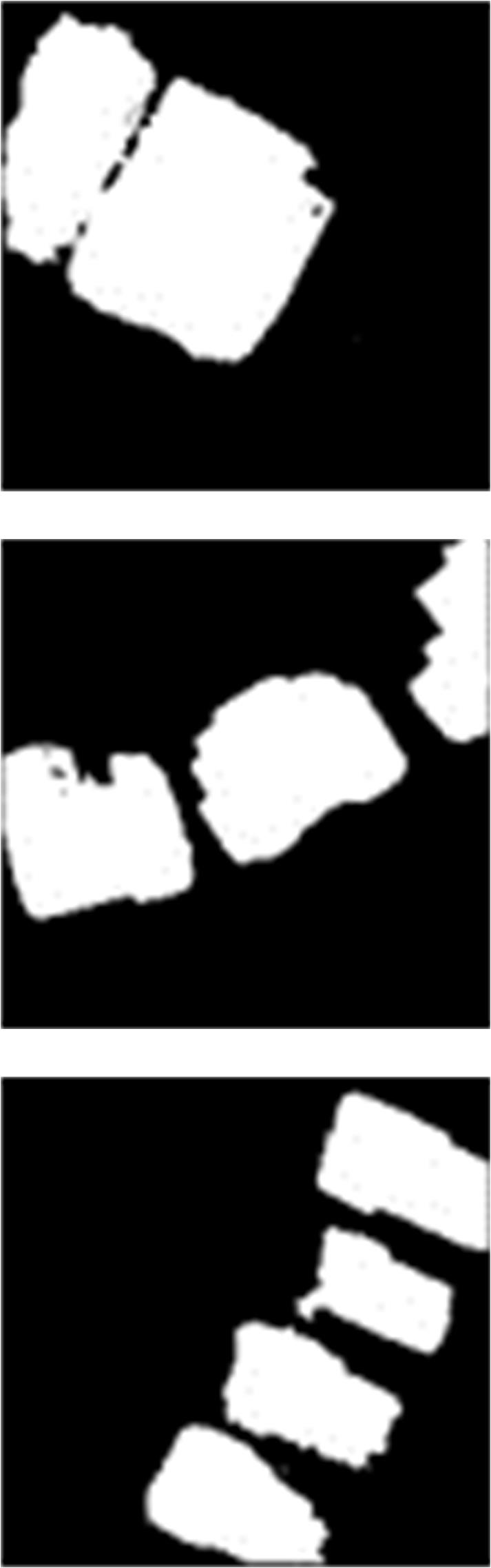}}
  \subfigure[]{
    \label{fig:subfig:g111} 
    \includegraphics[width=0.5in]{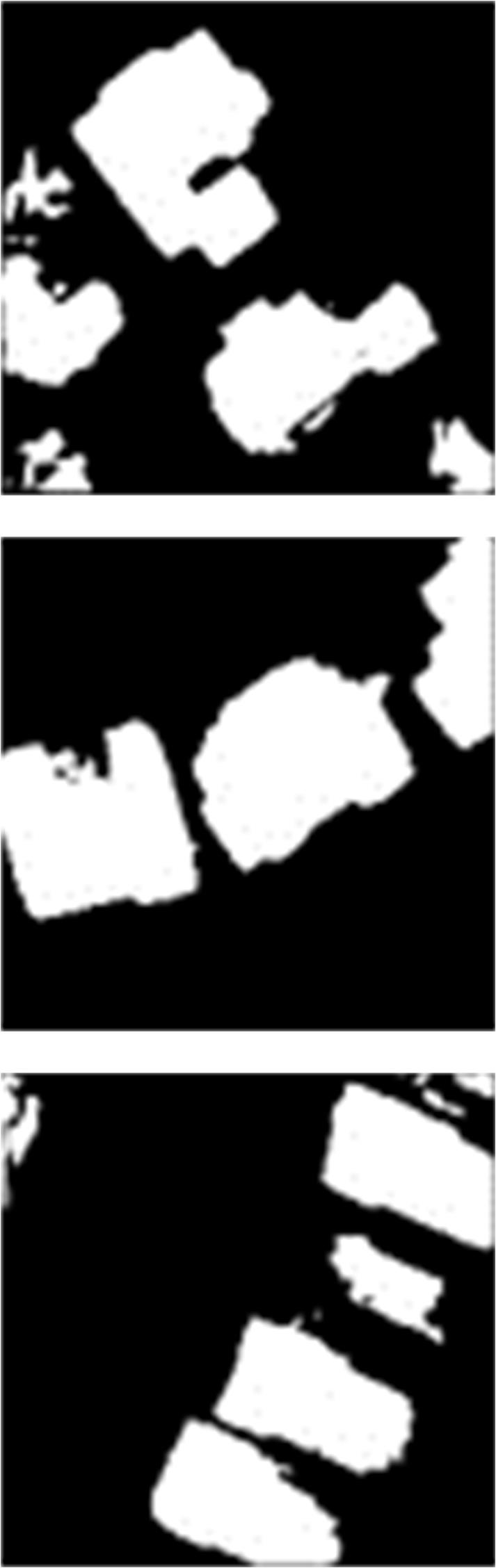}}
  \subfigure[]{
    \label{fig:subfig:h111} 
    \includegraphics[width=0.5in]{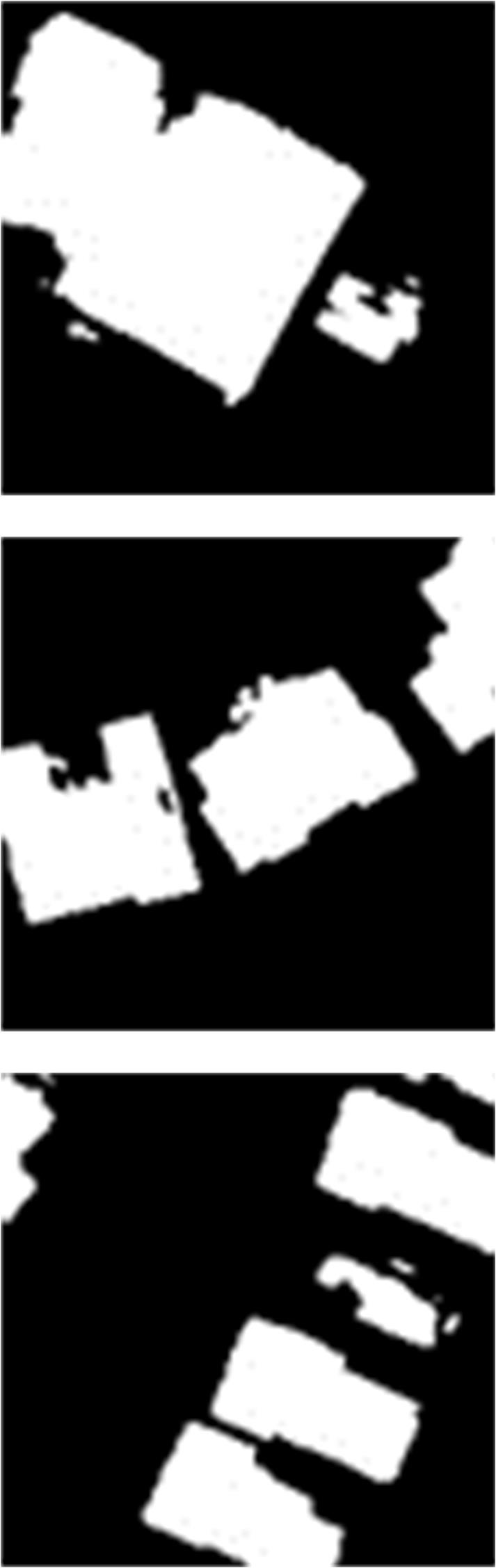}}
  \subfigure[]{
    \label{fig:subfig:i111} 
    \includegraphics[width=0.5in]{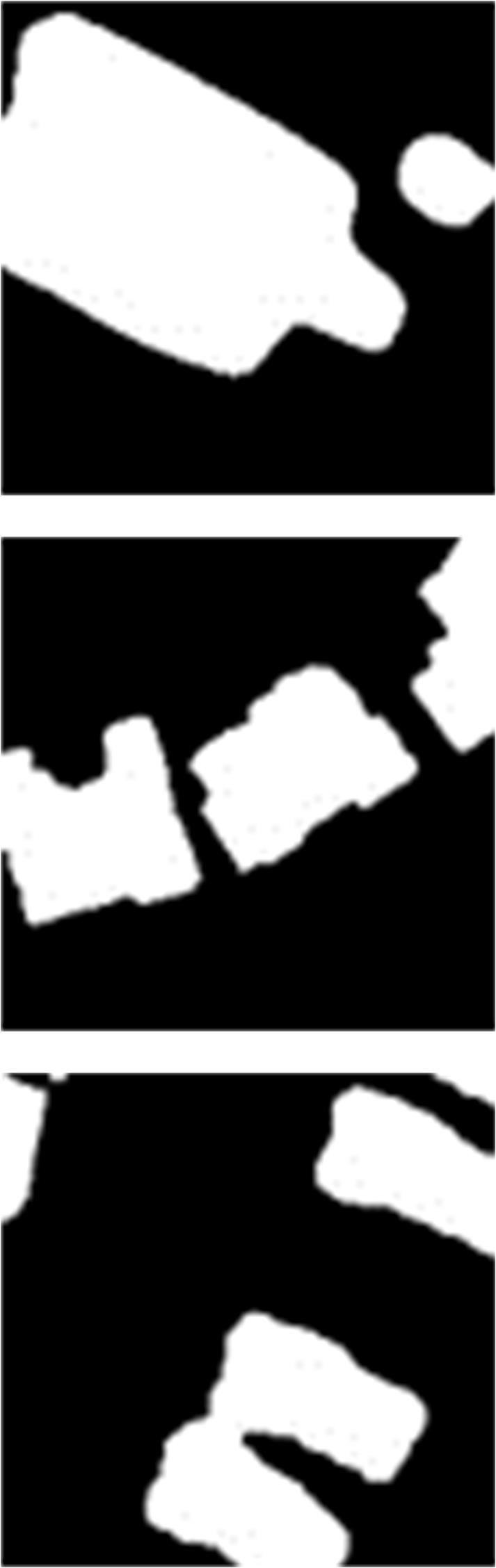}}
  \subfigure[]{
    \label{fig:subfig:j111} 
    \includegraphics[width=0.5in]{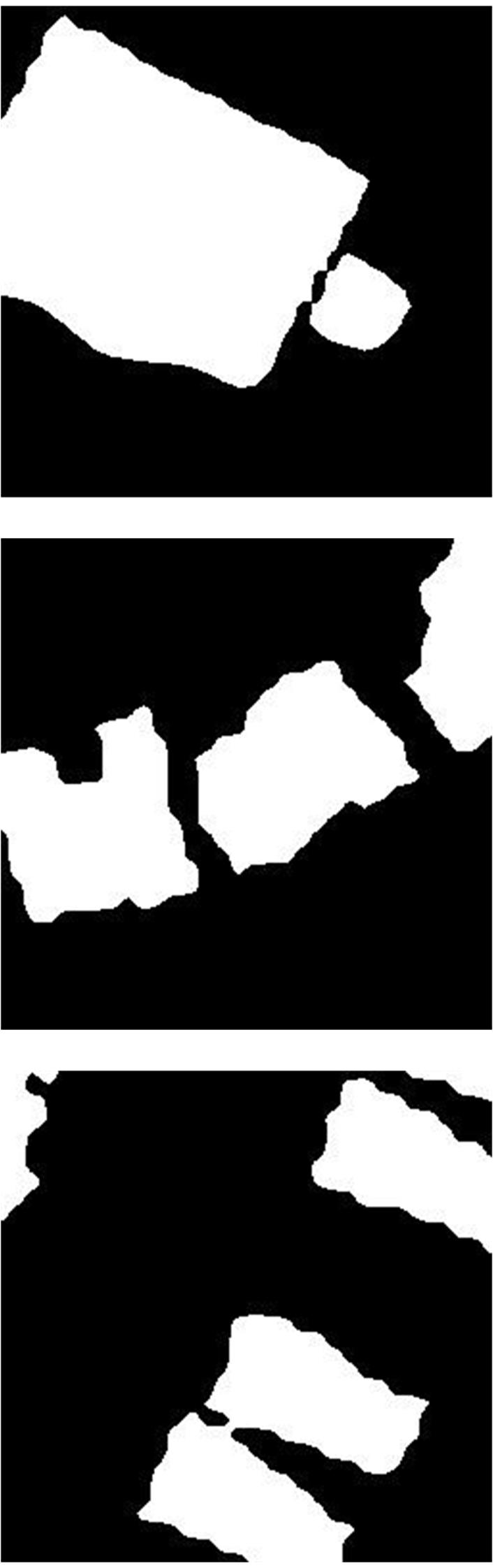}} 
  \caption{Visualized comparison of the results of various change detection methods on the BCDD dataset: (a) the unchanged image, (b) the changed image, (c) the label, (d) CDnet, (e) FC-EF, (f) FC-Siam-Diff, (g) FC-Siam-Con, (h) BiDateNet, (i) DASNet(VGG16), and (j) DASNet(ResNet50). The changed parts are shown in white, while the unchanged parts are shown in black.}
  \label{BCDDr} 

\end{center}
\end{figure*}

From the data in Table \ref{tab3} and Table \ref{tab4}, the proposed method performed significantly better than the other change detection methods. For a more intuitive evaluation, we visualized the experimental results (Fig.~\ref{CDDr} and Fig.~\ref{BCDDr}). According to Table 4, the proposed method realized satisfactory performance in remote sensing change detection. Compared with the other optimal change detection methods, the F1 score improved by 2.9\% on the CDD dataset and by 4.2\% on the BCDD dataset.

To evaluate the performance of our method more intuitively, we visualized the experimental results. According to Fig.~\ref{CDDr} and Fig.~\ref{BCDDr}, the proposed method realized satisfactory results on both the CDD and BCDD datasets.

In Fig.~\ref{CDDr}, the first row demonstrates our model's performance in recognizing small changes in complex scenarios. The colors of images before and after the change in the second row vary substantially, but this has a highly limited impact on our network; there are many pseudo-change interferences in the images before and after the change in the third row, while the change in the bottom-left corner of the fourth row is not readily observable. Our method performed well under both conditions.

According to Fig.~\ref{BCDDr}, compared with other change detection methods, the proposed method focuses more on the changing area of interest. Since the BCDD is a dataset for building change detection, it is only necessary to identify building changes. Other changes, such as road changes, can be regarded as pseudo-changes. The first row and the third row clearly show that the method proposed in this paper not only performs well in identifying changes but also exhibits satisfactory resistance to the interference of pseudo-changes.

\section{Conclusions and Future Work}
In this paper, we proposed a DASNet for high-score remote sensing image change detection, which directly measures changes by learning implicit metrics. To minimize the distance between the unchanged areas and maximize the distance between the changed areas, we used spatial attention and channel attention to obtain better feature representations and used the WDMC loss to balance the influences of the changed regions and the unchanged regions on the network. Compared with other baseline methods, our proposed network performed well on both the CDD and BCDD datasets. The Siamese network structure can learn the change representations of remote sensing images well, and the attention mechanism can describe the local features of the changes and recognize the pseudo-changes.

In the future, we will conduct further research on small samples and on open-world and noisy environments to improve the mobility and robustness of change detection.

\section*{Acknowledgment}

The authors would like to thank the people who share data and basic models of their research to the community. Their selflessness truly drives the research process for the entire community.


%





\ifCLASSOPTIONcaptionsoff
  \newpage
\fi





\bibliographystyle{IEEEtran}
\bibliography{Bibliography}{}

\vfill


\end{document}